\documentclass{article} 
\usepackage[final]{colm2026_conference}

\usepackage{multirow}
\usepackage{microtype}
\usepackage{hyperref}
\usepackage{url}
\usepackage{booktabs}
\usepackage{enumitem}
\usepackage{wrapfig}

\usepackage{lineno}

\hypersetup{colorlinks=true, citecolor=darkblue, linkcolor=darkblue, urlcolor=darkblue}

\usepackage{pifont}
\usepackage{makecell}

\definecolor{babyblue}{HTML}{4285f4}
\usepackage[most]{tcolorbox}
\usepackage{multirow}
\usepackage{bbm}
\usepackage{amsmath}

\usepackage{graphicx}
\usepackage{mathtools} 
\usepackage{algorithm}
\usepackage{algpseudocode}
\usepackage[svgnames,table]{xcolor}
\usepackage{xspace}
\newcommand{\sq}{\emph{Simple Questions}\xspace}
\newcommand{\wc}{\emph{NB-WildChat}\xspace}
\newcommand{\ic}{\emph{Infinity-Chat}\xspace}
\newcommand{\nb}{\emph{NB-Curated}\xspace}
\usepackage{cleveref}
\usepackage{csquotes}
\usepackage{array}
\usepackage{fontawesome5} 
\definecolor{darkblue}{rgb}{0, 0, 0.5}
\definecolor{infinitecol}{RGB}{224,219,205}
\definecolor{finitecol}{RGB}{218,231,227}
\newcommand{\faHuggingFace}{%
  \raisebox{-1.5pt}{\includegraphics[height=1.05em]{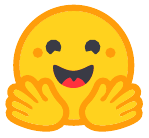}}\xspace}

\title{No Single Best Model for Diversity: \\ Learning a Router for Sample Diversity}

\newcommand{\aspace}{\hspace{1em}}
\author{%
    Yuhan Liu$^{\spadesuit}$ \aspace
    Fangyuan Xu$^{\spadesuit}$ \aspace
    Vishakh Padmakumar$^{\heartsuit}$ \\
    \textbf{Daphne Ippolito$^{\diamondsuit}$ \aspace
    Eunsol Choi$^{\spadesuit}$} \\
    $^{\spadesuit}$ New York University \aspace
    $^{\heartsuit}$ Stanford University \aspace
    $^{\diamondsuit}$ Carnegie Mellon University \\
    \texttt{yl13579@nyu.edu} \\
     \faGlobe~\href{https://diversity-router.github.io/}{diversity-router.github.io} \aspace
    \faGithub~\texttt{\href{https://github.com/lyh6560new/diversity_router}{diversity\_router}} \aspace
    \faHuggingFace\texttt{\href{https://huggingface.co/datasets/yuhan-nlp/diversity-router-data}{yuhan-nlp/diversity-router-data}}
}

\begin{document}

\ifcolmsubmission
\linenumbers
\fi

\maketitle

\begin{abstract}
When posed with prompts that permit a large number of valid answers, comprehensively generating them is the first step towards satisfying a wide range of users. In this paper, we study methods to elicit a comprehensive set of valid responses.
To evaluate this, we introduce \textbf{diversity coverage}, a metric that measures the total quality scores assigned to each \textbf{unique} answer in the predicted answer set relative to the  best possible answer set with the same number of answers.
Using this metric, we evaluate 18 LLMs, finding no single model dominates at generating diverse responses to a wide range of open-ended prompts. Yet, per each prompt, there exists a model that outperforms all other models significantly at generating a diverse answer set.
Motivated by this finding, we introduce a router that predicts the best model for each query. On \wc, our trained router outperforms the single best model baseline ($26.3\%$ vs $23.8\%$). We further show generalization to an out-of-domain dataset (\nb) as well as different answer-generation prompting strategies. Our work lays foundation for studying generating comprehensive answers when we have access to a suite of models.


\end{abstract}

\section{Introduction}

Various tasks require language models (LMs) to generate diverse high-quality responses. 
These range from creative writing~\citep{padmakumar2023does,chung2025modifying}, dialogues \citep{lintomlin2025usersim}, code \citep{wu2026x}, math \citep{wu2025mode}, scientific discovery \citep{novikov2025alphaevolve,gottweis2025towards}, survey response simulation \citep{meister2024benchmarking} to synthetic data generation \citep{honovich2022unnatural}.  Evaluation on these tasks thus should move beyond the quality of a single output to the diversity and quality of \textit{a set} of outputs. However, existing metrics fall short in quantifying the coverage of open-ended answer space.  In this work, we introduce \emph{diversity coverage}, a metric that measures the total quality scores assigned to all {unique} answers in the predicted answer set, relative to the best possible answer set with the same number of answers.

Prior works focused on methods for improving the diversity of generations from a single LLM, such as changing the inference hyperparameters~\citep {holtzman2019curious,kambhatla2022surfacing, santurkar2023whose,nguyen2024turning} or prompt \citep{lu2024benchmarking,zhang2025verbalized}.
In this paper, we ask the question of whether \emph{diversity coverage} can be further improved by taking advantage of bountiful LMs, each with differing behaviours towards the same prompt.
We hypothesize that heterogeneous LLMs can be effectively ensembled in order to leverage their complementary strengths.
Through a pilot study, we identify that no single LLM dominates in generating a diverse and high-quality output set, and different LLMs excel at answering different open-ended questions (\Cref{sec:pilot}). If we can pick the optimal model for each example, $33.0\%$ diversity coverage can be achieved on \wc, revealing a large gap compared to using the overall best model in the ensemble ($23.8\%$). 

\begin{figure}
    \centering
    \includegraphics[width=0.95\linewidth]{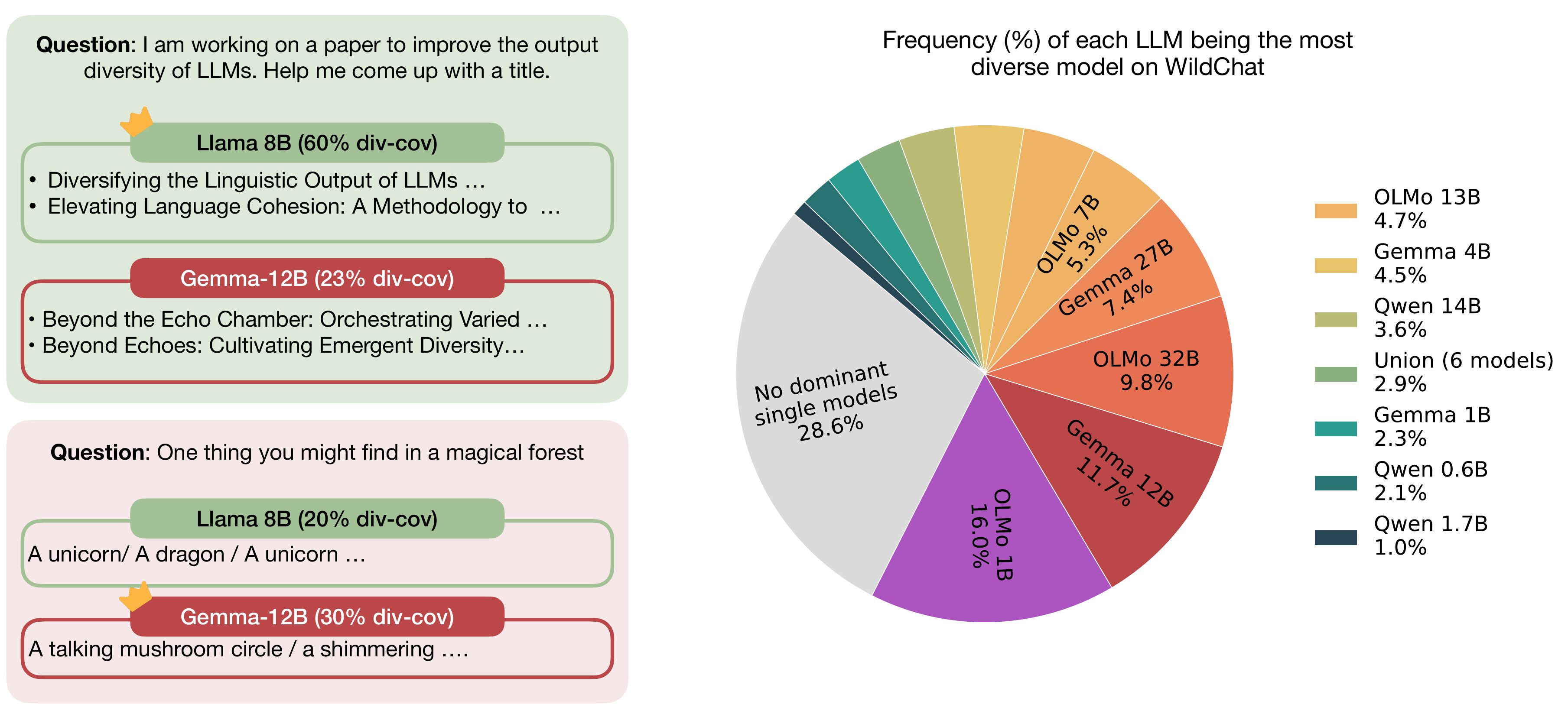}
    \caption{
    \textbf{Left}: LLMs exhibit different diversity coverage. \textbf{Right}: There is no universal best model on \wc. A model is only considered to be the best model if its diversity scores are $5\%$ higher than the second-best candidate. Queries without a model satisfying this margin are labeled as ``No dominant single models''. On \sq, all models perform similarly, resulting in $100\%$ of ``No dominant single models''. On \wc, there is no model that consistently dominates all queries.  In Appendix~\ref{app:dominant}, we take a look on frontier versions of GPT, Claude and Gemini and find no dominant LLM either.
    }
    \label{fig:teaser}
\end{figure}

However, determining the optimal model for a question can be challenging,  as optimizing diversity involves analyzing the joint effect of all sampled answers. Although many previous works have proposed routing to select the best LLM \citep{Jiang2023LLMBlenderEL,zhang2026router,lu-etal-2024-routing}, candidates are ranked based on a single response to the ground truth. This problem becomes exponentially more challenging when comparing answer sets as it is expensive to sample and compare multiple outputs for each example. The difficulty is further increased for open-ended questions as there are no gold answer sets available. Lastly, one might expect that simple heuristics based on model metadata (e.g., family or size) could help predict diversity. In fact, we show in Figure~\ref{fig:teaser} that models of different families and sizes generate the most diverse outputs for a disjoint set of queries.

Motivated by this, we propose a simple approach to predict the best model to respond to any given query.
We train a router that scores the diversity and quality of each candidate model in the routing pool and routes to the best LLM to generate the answers. We frame it as a classification task and create a training dataset of queries and model scores. Our router outperforms top overall baselines on the \wc and also generalizes to \nb. Using the same trained router, we explore further ensembling outputs from two models per query, which brings further performance gains. 
Our key contributions are:


\begin{enumerate}[leftmargin=*]
    \item We propose a new metric, \textit{diversity coverage}, to jointly measure the diversity and quality of an answer set to open-ended questions with diverse output space.
    \item Motivated by the finding that no single model excels at generating diverse outputs to all queries, we propose to train a router which predicts the best model for generation given an input query. We evaluate our method comprehensively on multiple datasets, demonstrating competitive performance on both in-domain (\wc) and out-of-domain (\nb) settings. 
    \item We further investigate the effect of training data size, prompting strategies as well as inference efficiency of using a diversity router.  More broadly, our findings highlight the potential of multi-LLM systems, where models with complementary strengths are combined to produce more diverse and high-quality solutions.
\end{enumerate}


\section{Task Formulation}\label{sec:bg}


Many queries admit multiple valid responses rather than a single correct answer. We summarize datasets containing such queries in Table \ref{tab:data}. Depending on the number of possible answers, we categorize the datasets into two types:
\begin{itemize}[noitemsep,leftmargin=10px]
    \item \textbf{Fixed answer set.} Each query has a finite ground truth answer set $A^*$. The predicted answer is correct if it belongs to $A^*$. We match model outputs to $A^*$ using exact string match.
    \item \textbf{Open-ended answer set.} Queries admit infinitely many valid answers, and listing them comprehensively is infeasible. The validity of the predicted answer can be evaluated based on quality and assigned a scalar value. We use a semantic equivalence classifier to determine whether two generated answers preserve the same meaning. 
\end{itemize}


\begin{table}
\footnotesize
    \centering
    \setlength{\aboverulesep}{0pt}
    \setlength{\belowrulesep}{0pt}
    \newcolumntype{L}[1]{>{\raggedright\arraybackslash}m{#1}}
   \begin{tabular}{L{3cm} c L{3.5cm} L{5cm}}
    \toprule
         \multirow{2}{*}{\textbf{Dataset}} &\multirow{2}{*}{\textbf{\# Q}} &\multicolumn{2}{c}{\textbf{Example}}\\
         \cmidrule(lr){3-4}
    & & \makecell[c]{Question} & \makecell[c]{Possible Valid Answers \\(delimited with `/')}\\ \midrule
         \rowcolor{finitecol}
         \sq{} \citep{zhang2024forcing} & $23$ & Output a random country in North America. &
         United States/ Canada/ Mexico / ...  \\ \midrule
         \rowcolor{infinitecol}
         \nb\footnotemark{} \citep{zhang2025noveltybench} & $92$ & Tell me a funny joke. &
         Who is Adam and why he is optimizing my code? / “Pick a lane” “Who’s Elaine?” \\
         \rowcolor{infinitecol}
         \wc{} \citep{zhang2025noveltybench} & $1$k & Give me a very simple way to remember the formula for tangent. &
         Tangent = $\frac{\text{Opposite}}{\text{Adjacent}}$ / Draw a unit circle ... \\
         \rowcolor{infinitecol}
         \ic{} \citep{jiang2026artificial} & $26$k & Write a story about America. &
         "... I'm just like my country—I'm young, scrappy, and hungry, and I'm not throwin' away my shot!" said Hamilton./ America was born in the streets. \\
         \bottomrule
    \end{tabular}\vspace{-0.4em}
   \caption{Dataset statistics. \# Q denotes the number of total questions. \colorbox{finitecol}{\strut\sq} has a fixed, finite answer set; all \colorbox{infinitecol}{\strut other datasets} have open-ended answer sets.}
    \label{tab:data}
\end{table}


\footnotetext{The original released dataset has 100 questions. We filter out 4 questions that ask for multiple answers (violating our hypothesis) and 4 questions that do not have multiple correct answers.}


\subsection{Task definition} 
Given a query $q$ and generation budget $B$ (i.e., the number of answers produced by model), the task is to derive an answer set $A =\{a_1, \ldots, a_B\}$ that covers as many distinct and high-quality answers as possible. Our task definition assumes two functions:


\begin{itemize}[noitemsep,leftmargin=10px]
    \item \textbf{uniq($q, A$)}: Given a query $q$ and an answer set $A$, it outputs a subset of $A_d$, consisting of distinct answers only (i.e., no two answers in $A_{d}$ are equivalent to each other). We follow prior work~\citep{zhang2025noveltybench} to derive $A_{d}$. We iterate over answers in $A$ and greedily add a new answer to $A_{d}$ if the current answer $a$ is not equivalent to any answer already in $A_{d}$. The equivalence of two answers is determined by exact string match for queries with a fixed answer set and an equivalence classifier for open-ended questions.\footnote{The classifier from \citet{zhang2025noveltybench} achieves $79\%$ accuracy and $0.811$ F1 against human labels on their held-out test set. We additionally evaluate it on 100 manually annotated \ic answer pairs and observe $88\%$ accuracy and $0.735$ F1.}
    The process is described in Appendix \ref{app:calc_details}. 
    
    \item \textbf{quality($q, a_i$)}: Given an individual answer $a_i$ for query $q$, it outputs a scalar value representing the quality of the answer $a_i$. For fixed-answer queries, this is computed by comparing against the ground-truth answer set; for open-ended queries, we use \texttt{Skywork-Reward-Gemma-2-27B-v0.2}~\citep{liu2024skywork} as the reward model, following \citet{zhang2025noveltybench}. Details are provided in Appendix~\ref{app:calc_details}.
    
    
\end{itemize}



\subsection{New Evaluation Metric: Diversity Coverage}




Given a predicted set of answers $A = \{a_1, \ldots, a_B\}$ to query $q$, 
we introduce a new metric, \emph{diversity coverage (div-cov)} as follows:
\[\operatorname{div-cov}(q,A)
\coloneqq
\frac{1}{\operatorname{max-uniq-sum}(q,B)}
\sum_{a \in \text{uniq}({q,A})} \operatorname{quality}(q,a)
\]  
We define \textbf{max-uniq-sum($q,B$)} as the maximum score that one can reach by generating an answer set of size $B$ where each answer in the set is distinct and achieves maximum quality.




For questions with a fixed answer set $A^*$, assuming $B \ge |A^*|$, this measures the proportion of unique ground-truth answers covered by the answer set, equivalent to the {coverage rate} metric proposed by \citet{zhang2024forcing}.




Prior works measure diversity by the number of unique, valid outputs for questions with a fixed answer set~\citep{zhang2024forcing}; or via pairwise embedding-based similarity 
\citep{zhang2025verbalized, jiang2026artificial}. Such metrics either do not work for questions with an open-ended answer space, or do not account for the quality of the answers.
\citet{zhang2025noveltybench} proposes a unified metric for quality and diversity considering an \textit{ordered} answer list, which penalizes answers generated later to account for user patience. This paper focuses on evaluating the quality and diversity of a set of answers, regardless of generation order, making diversity coverage better suited to our purpose.

\section{A pilot study on ensembling models to maximize diversity coverage}
\label{sec:pilot}

In this section, we first study whether a strong LLM can dominate other models in diversity coverage for a range of questions (Section~\ref{sec:pilot}). We found no single model dominates, motivating us to explore the upper bound of gains if we use a pool of LLMs instead of a single LLM (Section~\ref{subsec:oracle}). 
We compare several ensembling strategies under oracle model selection setting. We find that picking the best LLM per question is the most promising, which leads us to develop model router in later sections.

\subsection{No single model is best at diversity coverage for all questions}
\label{sec:no_single}



\paragraph{Model sets}
We study $18$ models from four open-source model families with different parameter counts: Llama (\texttt{Llama-3.2-1B}, \texttt{Llama-3.2-3B}, \texttt{Llama-3.1-8B}, \texttt{Llama-3.3-70B}), Qwen (\texttt{Qwen3-0.6B}, \texttt{Qwen3-1.7B}, \texttt{Qwen3-4B}, \texttt{Qwen3-8B}, \texttt{Qwen3-14B}, \texttt{Qwen2.5-72B}), OLMo (\texttt{OLMo-2-0425-1B}, \texttt{OLMo-2-1124-7B}, \texttt{OLMo-2-1124-13B}, \texttt{OLMo-2-0325-32B}), Gemma(\texttt{gemma-3-1b}, \texttt{gemma-3-4b}, \texttt{gemma-3-12b}, \texttt{gemma-3-27b}).

\paragraph{Settings}
For each model and query, we sample $N$ answers with a prompt which instructs the model to enumerate as many valid answers as possible (see Appendix \ref{app:prompts} for the full template). This prompt encourages models to explore the space of possible responses rather than produce a single canonical answer.\footnote{Prior work~\cite{zhang2025verbalized} has found that this method elicits more diverse answer set compared to sampling multiple single answer. We further compare different prompt templates (e.g. generate one or two answers in a single generation) in Appendix \ref{app:prompt_template}, finding that generating all answers in a single prompt elicits the most diverse answer set.} 
We keep the decoding method fixed throughout the paper (described in Appendix \ref{app:decoding}).
For each query, we define the ``dominant'' model by the following two criteria: (1) the model achieves the highest diversity coverage, and (2) the score is at least $5\%$ higher than that of the answer sets generated by any other models.  



\paragraph{Results} On \sq , we find no dominant model for any query. For \wc queries, this remains true for 30\% of queries, and more than 5 models are dominant at least on 5\% of queries, suggesting optimizing model choices per question can be fruitful. Figure \ref{fig:teaser} compares per-model frequency of achieving the best diversity coverage on \wc.

\subsection{Oracle experiment: how much does picking the best model(s) per query improve?}\label{subsec:oracle}
Motivated by the finding that different models can generate diverse outputs for different queries, can we \textit{ensemble} outputs from models to achieve diverse outputs? We assume an oracle setting, where we have access to the diversity coverage scores of all LLMs on all queries. 
We compare three strategies to ensemble models with a fixed generation budget $B$\footnote{In our experiments, we fix $B$ to be $50$ answers per question if not otherwise stated.}:

\begin{wraptable}{r}{0.6\textwidth}
    \vspace{-2em}
    \footnotesize
    \vskip 0.15in
    \centering
    \begin{tabular}{l c c c}
         \toprule
          & SQ & Curated & WildChat \\  
         \midrule

        Top overall model &$96.9\%$  &$47.0\%$ & $23.8\%$ \\ 
        Top two overall models  &$97.1\%$ &$45.6\%$ &$25.6\%$ \\
                Random model / query &$92.7\%$ &$37.5\%$ & $18.1\%$ \\
        Top model / query &$\textbf{97.9\%}$ &$\textbf{59.6\%}$ &$\textbf{33.0\%}$ \\
         %
         
         \bottomrule
    \end{tabular}
    \vspace{-0.5em}
    \caption[Caption]{Diversity coverage scores for ensembling multiple LLMs on \sq (SQ), \nb (Curated) and \wc (WildChat).
    }
    \label{tab:ensemble_gen_all}
\end{wraptable}

\begin{itemize}[leftmargin=10px]

    \item \textbf{Top overall model}. We select the single model with the best average diversity coverage per dataset, representing the best possible performance without ensembling. The selected top models are respectively: \texttt{Llama-3.1-8B}, \texttt{Qwen3-14B} and \texttt{OLMo-2-0425-1B}.


    \item \textbf{Top two overall models}. We select two models with the highest average diversity coverage score per the dataset, then ensemble their outputs, generating $B/2$ answers per model. The selected model pairs are respectively: (\texttt{Llama-3.1-8B}, \texttt{Llama-3.3-70B}), (\texttt{Qwen3-14B}, \texttt{Llama-3.1-8B}), (\texttt{OLMo-2-0425-1B}, \texttt{OLMo-2-1124-7B}).

    \item \textbf{Top model per query.} For \textit{each} query, we select the model with highest diversity coverage. 
    This represents the oracle performance of always choosing the best LLM per given query. We also report the performance of randomly choosing a model per query (\textbf{Random model per query}) as a baseline method.

\end{itemize}




We first show, in Table \ref{tab:ensemble_gen_all}, that model ensembling does improve output diversity, but the gains vary by task type: \sq sees near-saturated diversity coverage, with \emph{Top model per query} only slightly better than fixing a single model ($97.9\%$ vs. $96.9\%$). Open-ended questions show substantially larger gaps: on \nb, selecting the top model per query reaches $59.6\%$, a $27\%$ relative improvement over the best single-model baseline ($47.0\%$), and on \wc it improves over the best single-model baseline by $39\%$ ($33.0\%$ vs. $23.8\%$). We therefore focus the rest of our paper on open-ended datasets.
We then observe that query-level model selection, under oracle access, achieves the best diversity coverage. This motivates us to train a router for diversity in the next section.


\section{Learning to ensemble multiple models for diverse outputs} 
Oracle routing significantly improves diversity coverage, but it is costly as we need to sample and evaluate outputs from all candidate LLMs. This motivates us to train a router to predict the most promising model without sampling the entire answer sets from all models. 


\subsection{Router }
\paragraph{Problem setting}
Given a query $q$ and a suite of models $M = \{m_1, m_2,\cdots m_n\}$, a router ranks them, by $\operatorname{div-cov}(q,A_i)$ where $A_i$ is the generated answer set from $m_i$ for query $q$ for some budget $B$. The oracle model index for $q$ is defined as 
$i^* = \arg\max_{i} \operatorname{div-cov}(q, A^{(i)})$. Such index $i_j^*$  for each query $q_j$ consist of the router training data $\mathcal{D} = \{(q_j, i_j^*)\}$.





\paragraph{Classification Objectives} We compare two classification formulation for the router:
\begin{itemize}[leftmargin=10px]
   \item \textbf{$|\mathcal{M}|$-way classification:}
     the router is a single classifier
    $r_\theta:\mathcal{Q} \rightarrow \{1,\ldots,|\mathcal{M}|\}$ which predicts the oracle best model index $i_j^*$ for each query $q_j$. 
    Let $r_\theta(q)_i$ denote the predicted probability of selecting model $m_i$,
    we train the router with cross-entropy loss:
    $\mathcal{L}_{\text{multi}} =
    \mathbb{E}_{(q_j,i_j^*) \sim \mathcal{D}}
    \left[
    -\log r_\theta(q_j)_{i_j^*}
    \right].$
    \item \textbf{Binary classification:} For each LLM $m_i$, we derive a binary training dataset 
    $\mathcal{D}^{(i)} = \{(q_j, y_j^{(i)})\}$ from $\mathcal{D}$, where 
    $y_j^{(i)} = \mathbbm{1}[i = i_j^*]$ indicates whether $m_i$ is the oracle best model for query $q_j$. 
    We then train a binary classifier 
    $r_\theta^{(i)}:\mathcal{Q} \rightarrow [0,1]$ 
    to predict this label using binary cross-entropy loss.
    At inference time, the router evaluates all binary classifiers $\{r_\theta^{(i)}(q)\}_{i=1}^{|\mathcal{M}|}$ and selects the model with the highest predicted score:
$\arg\max_{i=1}^{|\mathcal{M}|} r_\theta^{(i)}(q)$.
\end{itemize}






\paragraph{Query encoding}
We experiment with two input featurizations:
(1) \textbf{infly/inf-retriever-v1}~\citep{infly-ai_2025}, a retriever fine-tuned from \texttt{Qwen-2-7B} for information retrieval tasks. We refer to it as model-agnostic encodings (agn).
(2) \textbf{Model hidden states}: we encode the query using each model $m_i$ and extract the representation from the final layer’s last hidden state. We hypothesize that this representation encodes rich information on how the model decodes its outputs. We refer to it as model-specific encodings (spec).





\subsection{Experiment settings}

\paragraph{Training and evaluation data}
We split the $1,000$ \wc~ prompts from \citep{zhang2025noveltybench} into train, validation and test sets containing 70\%, 10\% and 20\% of the data respectively. We conduct out-of-domain evaluation on \nb questions.  




\paragraph{Evaluation metrics}
Diversity coverage jointly measures the diversity and quality of the generated answer set.
To disentangle the effect, we additionally report metrics that measure each aspect.
\textbf{Quality (Qual)} measures the average quality score across all sampled answers:
$\frac{1}{|A|}\sum_{a\in A}\operatorname{quality}(q,a).$
\textbf{Uniqueness (Unq)} measures the number of semantically non-equivalent answers:
$|A_d|$.
\textbf{Unique Quality (Unq Qual)} measures the average quality score over unique answers only:
$\frac{1}{|A_d|}\sum_{a\in {\text{uniq}(q, A)} }\operatorname{quality}(q,a).$ Together, these metrics reveal whether improvements in cumulative diversity arise from generating more distinct answers, improving answer quality, or both.

\paragraph{Baselines} We consider several \textbf{non-routing baselines}. For a fair comparison with our trained routers, we restrict these methods to access only training-set labels and evaluate them on the test set.
We implement baselines from Section~\ref{subsec:oracle}: \emph{Top overall}, \emph{Top two overall}, \emph{Random model per query}. 
We also include a \emph{Frequency} baseline, where models are sampled proportional to their frequency of reaching highest diversity coverage. 
 We additionally compute \emph{Top model per query} and \emph{Top two models per query} as \textbf{oracle} performance on diversity coverage, using ground-truth labels on the test set. Specifically, \emph{Top model per query} is implemented by selecting the best model per query. 
\emph{Top two models per query} are the best pair over all model combinations. If two models are selected, we take half from each model.  


%


\paragraph{Router Models} We implement three types of router models, and describe their implementation details in Appendix \ref{app:router_implementation}. 
\begin{itemize}[leftmargin=12px]
    \item \textbf{KNN \citep{fix1985discriminatory}} This is a simple, non-parametric classifier, where the predictions are obtained from K nearest neighbours from the training data, $K\in {1,5}$. 
    \item \textbf{BERT \citep{devlin2019bert}} Following other routing literature~\citep{ong2025routellm,zhang2026router}, we fine-tune BERT with a classification head which makes a selection over the $|\mathcal{M}|$ models.\footnote{We did not experiment with implementing $|\mathcal{M}|$ BERT models for binary classification given that fine-tuning BERT is computationally more expensive than fine-tuning the 2-layer MLP router.} 
    \item \textbf{MLP} We  report results for training $|\mathcal{M}|$ binary MLP classifiers and training one MLP classifier for  $|\mathcal{M}|$-way classification. We report results for (1) using {inf-retriever} to encode the query for all classifiers: Binary MLP (agn) and  M-way MLP (agn), (2) using the candidate model $m_{i}$'s last layer hidden states to encode the query for the respective classifier: Binary MLP (spec) and M-way MLP (spec). 
    
\end{itemize}

\begin{table*}[t]
    \footnotesize
    \centering
    {\setlength{\tabcolsep}{3.5pt}
    \begin{tabular}{lcccccccc}
    \toprule
        \multirow{2}{*}{Method} & \multicolumn{4}{c}{NB-WildChat} & \multicolumn{4}{c}{NB-Curated (OOD)}  \\
        \cmidrule(lr){2-5} \cmidrule(lr){6-9}
          &\#Unq &Qual &Unq Qual &Cov.   &\#Unq &Qual &Unq Qual &Cov.    \\ \midrule
          Top overall   &42.6 &3.0 &2.9 &{23.8\%}  &35.4 &6.0 &5.7 &{38.6\%} \\ 
         Frequency  &33.1 &3.8 &3.6 &21.0\%   &28.2 &7.2 &7.1 &39.6\%  \\
         Random model per query    &27.8 &3.7 &3.6 &18.1\%   &27.8 &7.2 &7.0 &37.5\% \\
         Top model per query \small (oracle)     & 38.8 &4.5 &4.4 &33.0\%  &30.3 &7.6 &7.4 &59.6\%  \\ \midrule
         KNN (N=1)     &34.3 &3.7 &3.6 &23.1\%   &28.2 &7.3 &7.1 &39.7\%  \\
         KNN (N=5)      &34.9 & 3.8 &3.7 &24.1\%  &29.8 &7.3 &7.1 &40.2\% \\
         M-way BERT &40.3 &3.3 &3.2 &24.4\%   &35.0 &6.3 &6.2 &40.3\%\\ 
         M-way MLP (agn) &35.1 &3.9 &3.8 &25.3\%  &30.1 &7.6 &7.5 &40.3\% \\
         M-way MLP (spec)    &39.3 &3.5 &3.4 &25.9\%  &34.6 &6.3 &6.1 &40.2\% \\
         Binary MLP (agn) &38.4 &3.5 &3.4 &25.7\%$^{**}$ &32.8 &7.1 &7.0 &\textbf{40.7\%}$^{**}$ \\
         Binary MLP (spec)    &38.1 &3.6 &3.5 &\textbf{26.3\%}$^{**}$ &30.8 &7.0 &6.8 &39.3\%$^{ns}$ \\ 
         \bottomrule
         
    \end{tabular}
    }
     \caption[Caption]{A per-query router selects among 18 models to maximize diversity coverage (Cov.). 
     We train our best MLP router for 5 runs with random seeds to compute statistical significance for our best system (\textbf{bolded}) against \emph{Top overall}, $^{**}$ indicating significantly better and $^{ns}$ indicating not significant.
     }
     \label{tab:router}
\end{table*}


\section{Results}

\subsection{Performance Evaluation}
We report performances in Table \ref{tab:router}.\footnote{We also report accuracy (i.e., how frequently it predicted ground truth best model) in Table \ref{tab:router_acc} in Appendix.} 
\emph{Top overall} is the best-performing non-routing baseline for in-domain evaluation on \wc. This indicates that the LLM chosen from training labels maintains strong diversity coverage on the test set. 
\emph{Frequency} baseline generalizes better to out-of-domain \nb questions.
KNN routers yield only marginal improvement. MLP-based routers outperform other baselines. Specifically, 
binary routers with model-specific query encodings bring the greatest gains ($26.3\%$), surpassing the \emph{Top overall} baseline ($23.8\%$). On MLP classifiers,  model-specific query encodings (spec) provide more useful information than model-agnostic encoding (agn), but show worse generalization. 

\begin{table*}[t]
    \footnotesize
    
    \setlength{\tabcolsep}{4pt}
    \centering
    \begin{tabular}{lcccccccc}
    \toprule
        \multirow{2}{*}{Method} & \multicolumn{4}{c}{NB-WildChat} & \multicolumn{4}{c}{NB-Curated (OOD)}  \\
        \cmidrule(lr){2-5} \cmidrule(lr){6-9}
          &\#Unq &Qual &Unq Qual &Cov.   &\#Unq &Qual &Unq Qual &Cov.   \\ \midrule
         Top 2 overall &39.1 &3.4&3.2 &23.8\% &31.6 &6.7 &6.3 &38.3\%\\ 
         Top 2 per query  &40.7 &4.5 &4.5 &35.8\% &41.3 &7.7 &7.6 &62.6\% \\ \midrule
        Router  &38.4 &3.8 &3.6 &\textbf{26.7\%}$^{**}$ &32.3 &7.1 &6.8 &\textbf{42.2\%}$^{**}$ \\
         \bottomrule
         
    \end{tabular}
    \caption[Caption]{Performance of ensembling two models per query. We report the performance of the best single model router (Binary MLP (spec)) in \autoref{tab:router} by ensembling the top 2 models ranked by the prediction scores. $^{**}$ indicating significantly better compared to \emph{Top 2 overall}.}
     \label{tab:router_two}
\end{table*}

\paragraph{Router trained to select single model can be used to ensemble outputs from two models which provides further gains.}
We observe consistent gains when using our trained router (Binary MLP(spec)) to select two models, as presented  in Table \ref{tab:router_two} both in-domain ($26.7\%$ vs. $23.8\%$) and out-of-domain ($42.2\%$ vs. $38.3\%)$. Moving from \emph{Top overall} to \emph{Top 2 overall}, the best diversity coverage of the non-routing baseline does not improve. But the oracle (\emph{Top  per query}) stably increases from one to two models. We show that our best routers are significantly better than \emph{Top overall} and  \emph{Top two overall} baselines across 5 checkpoints trained under different random seeds. We further discuss how the number of models selected affect the answer diversity in Section \ref{app:configs}. 

\begin{wrapfigure}{r}{0.5\linewidth}
  \centering
  \vspace{-1.5em}
  \includegraphics[width=0.6\linewidth,
  trim=13 0 0 0,
  clip]{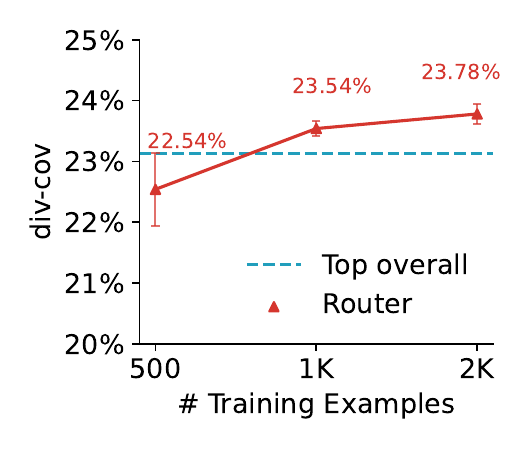}
  \vspace{-2em}
  \caption{Scaling training data improves router performance on \ic.}
  \label{fig:scaling_router}
  \vspace{-1em}
\end{wrapfigure}

\paragraph{Scaling training data size consistently produces a better router.} 
Would training on a larger data set improve performance? On \ic,  we show in Figure \ref{fig:scaling_router} that router performance increases steadily with training data sizes  varying from 500, 1k to 2k. We further find that training also scales on \wc and can incur generalization across the two datasets in Appendix \ref{app:scaling_router}.

\begin{wrapfigure}{r}{0.5\linewidth}
    \centering

    \vspace{-1.5em}
    \includegraphics[width=1\linewidth]{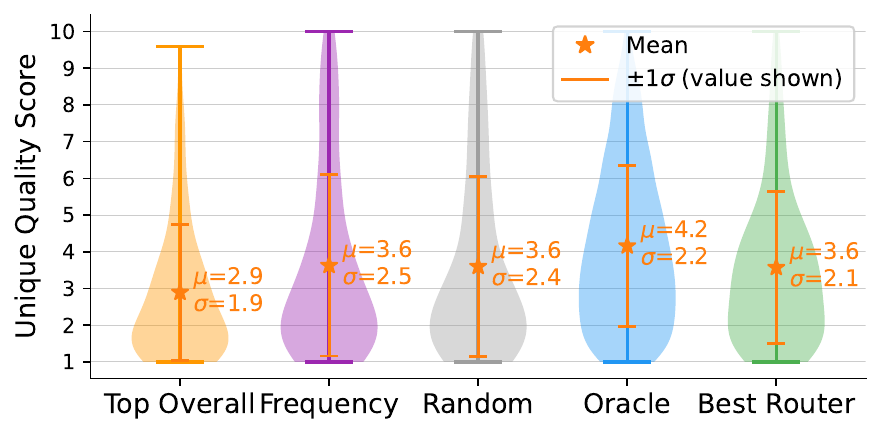}
    \vspace{-2em}
    \caption{Distribution of quality scores for unique answers on the WildChat dataset across different routing methods. The oracle router achieves the highest mean quality, while the learned router attains relatively high mean quality with low variability.}
    \label{fig:unique_quality_dist_router}
     \vspace{-1em}
\end{wrapfigure}
\paragraph{Answer quality distribution.}
Diversity coverage provides an aggregate measure of the quality and uniqueness of an answer set. We further examine how the quality scores of unique answers are distributed within each set. The oracle router achieves the highest mean quality ($\mu=4.2$) while maintaining moderate variability ($\sigma=2.2$). Among the non-oracle methods, the learned router achieves a mean quality of $3.6$, tying with the \textit{Frequency} and \textit{Random} baselines, while exhibiting the second-lowest standard deviation ($\sigma=2.1$). This suggests that the learned router produces answers with relatively high and consistent quality. In Appendix \ref{app:quality_variance}, we additionally include raw answer quality distribution and a comparison among different routers and different models.



\subsection{Efficiency Evaluation}

\begin{wrapfigure}{r}{0.6\linewidth}
    \centering

    \vspace{-2em}
    \includegraphics[width=1\linewidth]{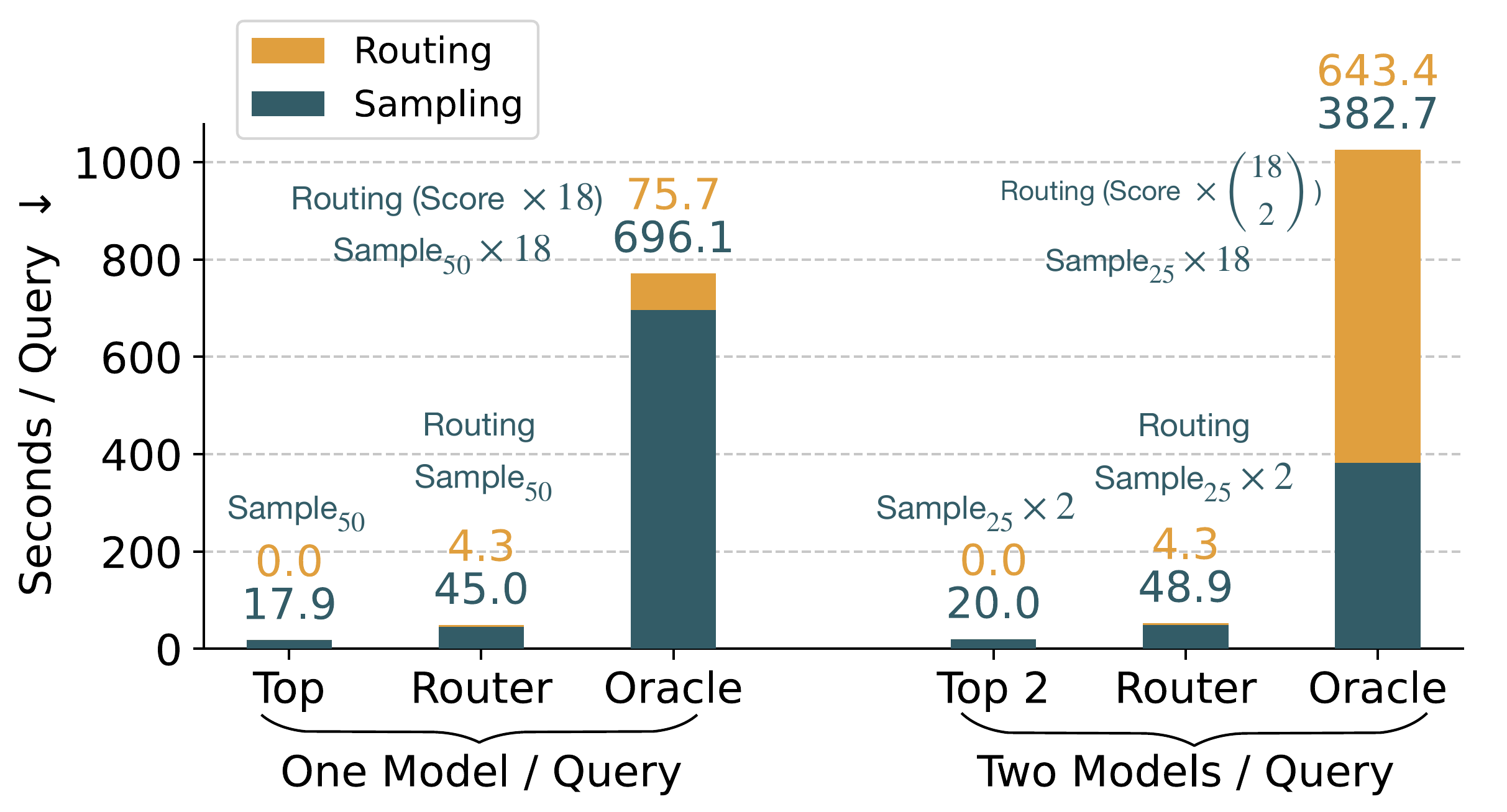}
    \vspace{-2em}
    \caption{Efficiency analysis comparing the time (seconds per query) across routing (Router), \emph{Top overall} (Top) and \emph{Top model per query} (Oracle).  We include routing to one model per query and routing to 2 models per query.
    Sample$_{n}$ denotes sample $n$ answers. Oracle incurs the highest cost, as the routing requires exhaustively comparing all candidate models or model pairs. 
    }
    \label{fig:efficiency}
     \vspace{-2em}
\end{wrapfigure}

In Figure \ref{fig:efficiency}, we show the inference time efficiency of generating an answer set of various methods on \wc. We use 2H200 GPUs for answer sampling\footnote{We assume no parallelization in sampling generations. If two models are selected, the process is performed sequentially (i.e. model by model). If multiple selected models are served in parallel, wall-clock latency may be lower than our sequential estimate, but this requires additional serving resources such as keeping multiple models loaded simultaneously. We leave a fuller analysis of parallel multi-model serving to future work.} and 1 H200 GPU for diversity coverage calculation. We compare the latency of three methods: \textbf{Top} (\emph{Top overall}), \textbf{Router},  which is the Binary MLP classifier (spec), and \textbf{Oracle} (\emph{Top model per query}).

Inferencing with our router is about $2\text{-}3\times$ slower than the Top baseline. The routing itself is not very costly, yet sampling becomes more expensive as the router often directs to a bigger LLM than the top baseline model (\texttt{OLMo-2-0425-1B}). 

Oracle setting, while showing the strongest performance, is also much more expensive, introducing up to $19\times$ computation overhead compared to our router. This is because its routing involves brute-force computing diversity coverage for all models to find the best candidate per query. 
In contrast, our router introduces only a fixed overhead that does not scale with the number of selected models. 

\section{Discussions: Different Prompt Templates}

\label{sec:future}

\begin{wraptable}{l}{0.5\linewidth}
\vspace{-1.4em}
 \footnotesize
     \centering
    \begin{tabular}{lccc}
    \toprule
  \multirow{2}{*}{Method} 
  &\multicolumn{3}{c}{Prompt Template} \\
  &\multicolumn{1}{c}{Gen 1} &\multicolumn{1}{c}{Gen 2} &\multicolumn{1}{c}{Gen All}\\
     \midrule
    Top overall &18.5\% &19.7\% &23.8\%\\
    Random &9.9\% &13.2\% &18.1\% \\
    Frequency &15.6\% &17.1\% &21.0\% \\ \midrule
    Oracle (G-1) & \cellcolor{LightGrey}25.6\% &\cellcolor{LightGrey!40}22.3\% &\cellcolor{LightGrey!40}20.4\%\\
    Oracle (G-2) &\cellcolor{LightGrey!40}19.5\% &\cellcolor{LightGrey}28.3\% &\cellcolor{LightGrey!40}21.0\% \\
    Oracle (G-All)  &\cellcolor{LightGrey!40}14.7\% &\cellcolor{LightGrey!40}18.8\% &\cellcolor{LightGrey}33.0\%  \\
    \midrule
    Router (G-1) &\cellcolor{LightGrey}19.1\% &\cellcolor{LightGrey!40}20.4\% &\cellcolor{LightGrey!40}14.4\% \\ 
    Router (G-2) &\cellcolor{LightGrey!40}\textbf{19.7\%} &\cellcolor{LightGrey}\textbf{21.6\%} &\cellcolor{LightGrey!40}19.0\%\\
    \makecell[l]{Router (G-All)} &\cellcolor{LightGrey!40}14.8\% &\cellcolor{LightGrey!40}18.1\% &\cellcolor{LightGrey}\textbf{26.3\%} \\
    \bottomrule
    \end{tabular}
    \caption[Caption]{Div-Cov (\%) results  on \wc with various prompting strategies. Training the router under each prompting strategy (\colorbox{LightGrey}{in domain} and \colorbox{LightGrey!40}{out-of-domain} evaluation).    }
     \label{tab:router_prompt}
     \vspace{-3.7em}
\end{wraptable}

Large amount of work in diversity has focused on improving the prompt, while throughout this paper we used a fixed prompt template to sample answers and compute diversity coverage. In this section, we explore two alternative prompt templates, with the exact prompts provided in Appendix~\ref{app:prompts}: 
\begin{itemize}[noitemsep,leftmargin=10px]
    \item \textbf{Generate one (G-1)}: to produce \textit{one} random answer for the given question.
    \item \textbf{Generate two (G-2)}: to provide \textit{two} different answers for given question.
    \item \textbf{Generate all (G-All)}: to list \textit{all} possible answers sequentially. This is our default.
\end{itemize}

Table~\ref{tab:router_prompt} summarizes the results. We use the same baselines as in Table~\ref{tab:router} in the first block. Comparing across three prompts, we find our default prompt (G-All) overall achieves the highest performance, as shown by the diversity scores (Cov.) in the first block.

In the second and third block, we report the oracle (\emph{Top model per query}) and router results. We use our best router (Binary MLP(spec)) for the experiment.
Router (X) is a router trained under prompt type X. Oracle (X) denotes that we always use ground truth labels derived by sampling with prompt X as predictions. Training a router improves diversity for all prompts, as all routers beat their \emph{Top overall} baselines. However, we see little generalization in both oracle and trained router across prompts. For instance, when generating with G-1 prompt, Oracle model chosen for the G-All prompt performs worse than baselines under G-1 prompt. Moreover, larger gains are observed when routing under better prompts.
You can find more detailed comparison in Appendix \ref{app:prompt_template}.

\begin{wrapfigure}{r}{0.55\linewidth}
    \centering
    \vspace{-1em}
    \includegraphics[width=1\linewidth]{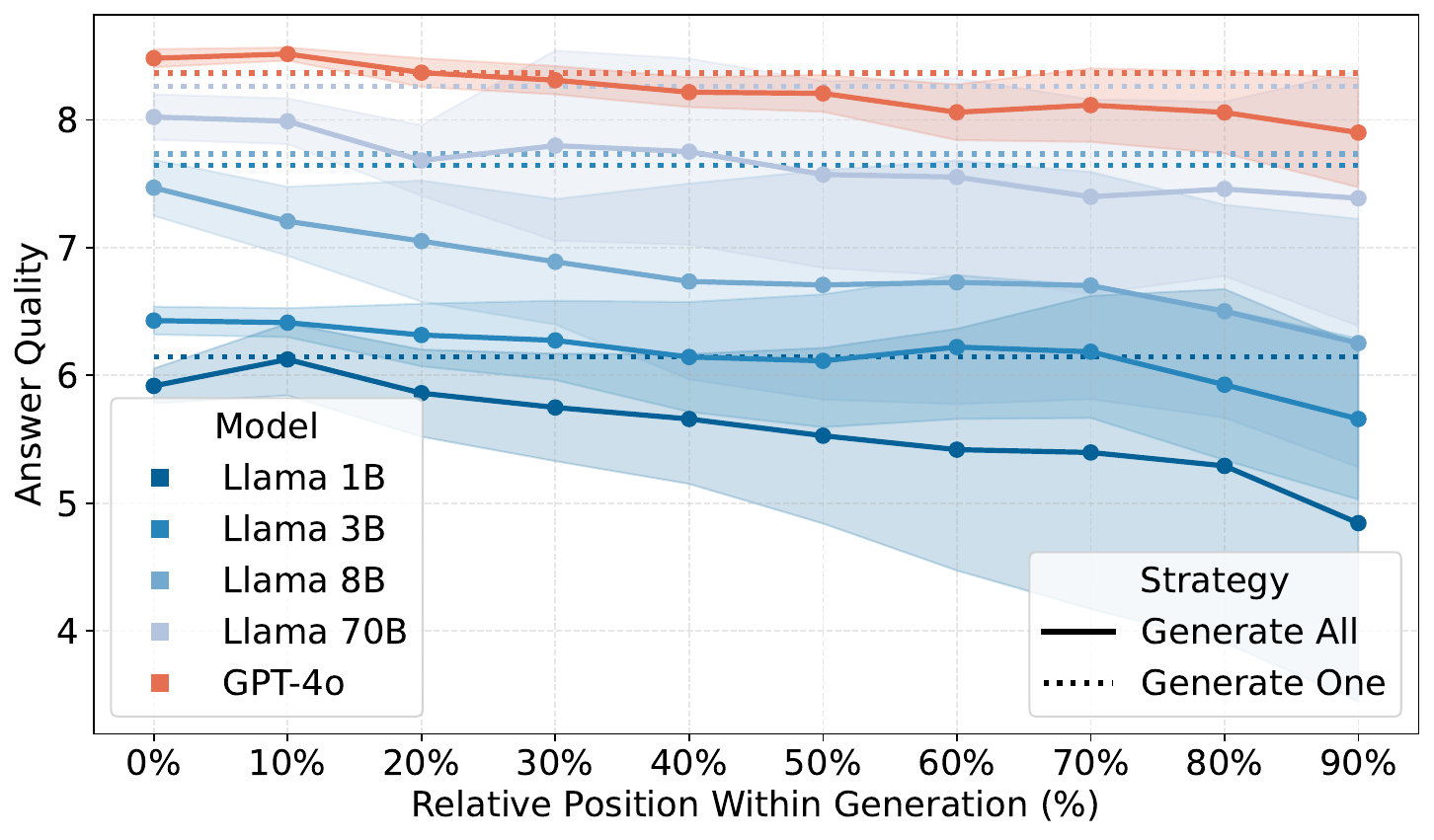}
    \vspace{-2em}
    \caption{Generate-one prompt has higher answer quality. Under the generate-all prompts, as more answers are listed,  the quality decreases with large variations if using generate-all prompt. 
    }
    \label{fig:score_position}
    \vspace{-1em}
\end{wrapfigure}

\paragraph{Degrading Answer Quality While Listing Multiple Answers}

Should we always use G-All prompt? 
Figure~\ref{fig:score_position} plots average answer quality under two prompt strategies (G-1 and G-All). For the generate-all prompt, we plot the quality of answers at different locations within the same generation. For the generate-one prompt, which produces one answer per generation, the quality is plotted as one dashed line.
We find two trends: (1) G-1 prompt consistently generates answers with higher average answer quality than G-All prompt and (2) In G-All prompt, as the generation continues, the answer quality decreases and the variance of quality scores increases. Therefore, when individual answer quality is more important, G-1 prompt, while harder to elicit diverse answers, can be more appropriate.

\section{Related Work}

\paragraph{Improving output diversity}

Concerns about the output diversity of LLMs \citep{padmakumar2023does, anderson2024homogenization, west2025base} promoted two categories of solutions: methods that modify model weights \citep{lanchantin2025diverse, chung2025modifying,sorensen2025spectrum, puri2026reachingmoderldistributional} and inference methods \citep{welleck2024decoding,levy-etal-2023-diverse, meister2024benchmarking, xiao2025role,kambhatla2022surfacing, santurkar2023whose,hayati-etal-2024-far,wang2025multilingual}. 
A suite of work proposes advanced prompting strategies, such as denial prompting \citep{lu2024benchmarking}, probabilistic prompting \citep{wong2024simplestrat}, and verbalized sampling \citep{zhang2025verbalized}. All of these methods focus on improving the diversity of a single model, whereas we study a multi-LLM setting.  MMR-style relevance-diversity reranking~\citep{goldstein-carbonell-1998-summarization} iteratively selects candidates that are both relevant to the query and dissimilar to already selected candidates. Such methods assume that a candidate pool has already been generated. In contrast, our router decides which model to sample from before generation under a fixed budget. Combining model routing with MMR-style reranking after generation is an interesting direction for future work.


\paragraph{Routers for LLMs} Researchers find that looping in multiple models is often better than sticking to one~\citep{Jiang2023LLMBlenderEL,feng-etal-2024-modular,feng2025llmdroolsmultillmcollaboration,feng2025heterogeneousswarmsjointlyoptimizing, feng2026mocoonestopshopmodel, feng2026singlemultievolutionloopselfimproving}.
Building on this insight, many works train a router that selects among multiple LLMs to achieve better task performance \citep{Jiang2023LLMBlenderEL,zhang2026router,lu-etal-2024-routing} or efficiency \citep{chen2024routerdc,ding2024hybrid,ong2025routellm,zhang2026router}. Simple methods \citep{ding2024hybrid,ong2025routellm} demonstrate the effectiveness of routing by switching between a stronger and a weaker model, which balances cost and quality.   Other works \citep{Jiang2023LLMBlenderEL,lu-etal-2024-routing} train routers with many top performing LLMs to leverage their complementary expertise.  
All existing methods are proposed to enhance the end performance measured within a single generation per question. 
However, none of the above discusses how routing can benefit the diversity and quality of a set of derived answers.
To the best of our knowledge, we are the first to propose a router to promote diversity coverage by harnessing the complementary efforts from heterogeneous models.


\section{Conclusion}
In this paper, we study mixing outputs from multiple LLMs as a strategy to improve response diversity. We first formalize diversity as the coverage of high-quality responses and propose unified evaluation metrics that apply to both finite and open-ended answer spaces. To optimize these metrics, we introduce a router that dynamically selects the most suitable LLM(s) for each query, showing improved performance. Further scaling the training data consistently improves the router.
We make a few simplifying assumptions: (1) when two models are selected, their outputs are mixed in equal proportions; and (2) only one or two models are used per query. Future research is encouraged to relax these limitations and explore efficiency-aware routing. 

\section*{Acknowledgments}
This work was supported in part through the NYU IT High Performance Computing resources, services, and staff expertise. The work is partially funded by NSF CAREER award 2443271.


\bibliography{colm2026_conference}
\bibliographystyle{colm2026_conference}

\newpage
\appendix

\section{The distribution of most diverse models}
\label{app:dominant}
\begin{figure}[h]
    \centering
    \includegraphics[width=0.7\linewidth]{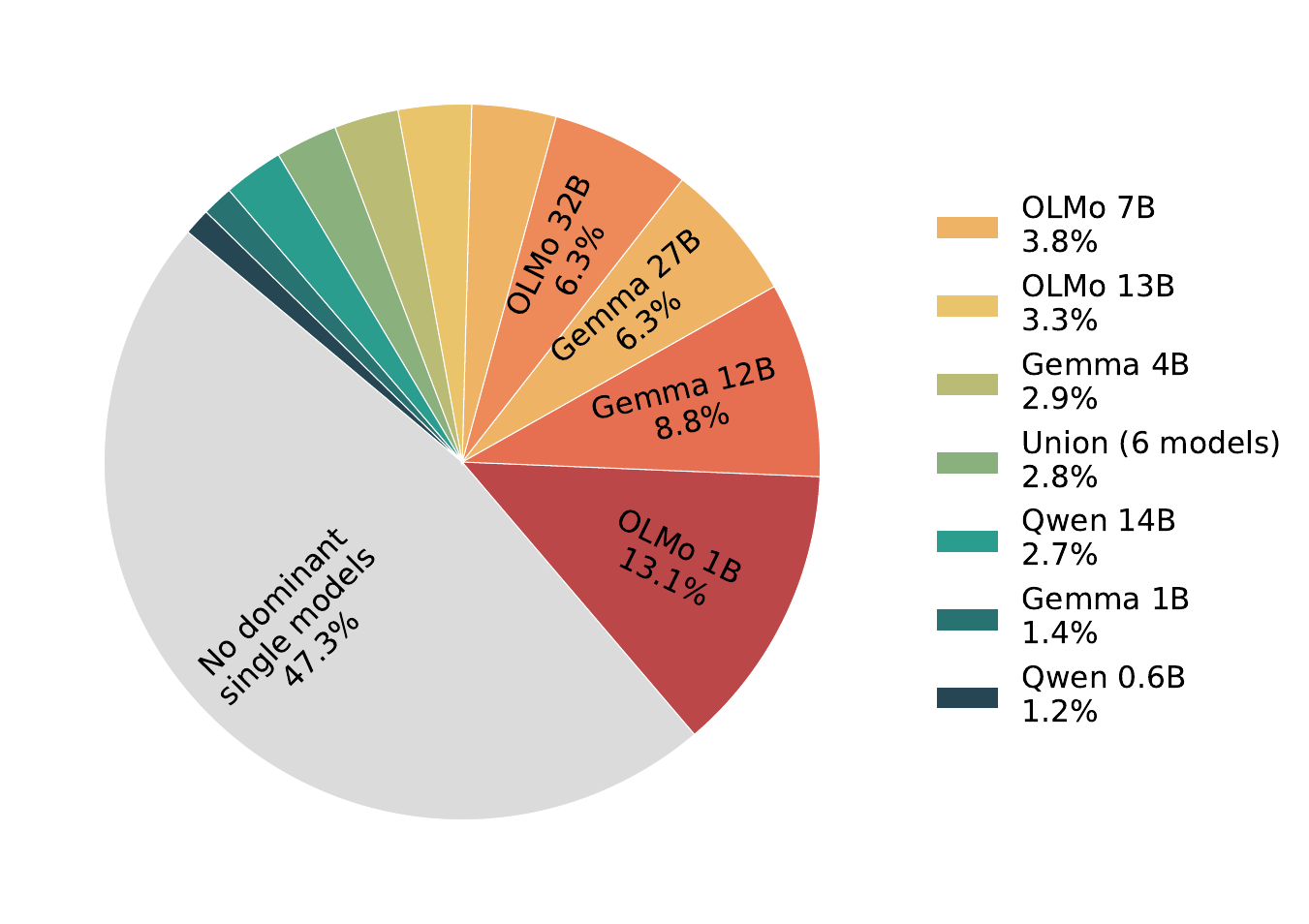}
    \caption{Frequency(\%) of each LLM being the most diverse model.A model is only considered to be the best model if its diversity coverage is $10\%$ higher than the second  best candidate. Queries with no meaningful gap are labeled as ``No dominant single models''. On \wc, there is no model that consistently dominates all queries.  On \sq,  all models have similar diversity coverage.
    }
    \label{fig:teaser_0.1}
\end{figure}

We attach in Figure \ref{fig:teaser_0.1} the frequency of each model being the best model if the threshold set to $10\%$.

\begin{figure}[h]
    \centering
    \includegraphics[width=0.5\linewidth]{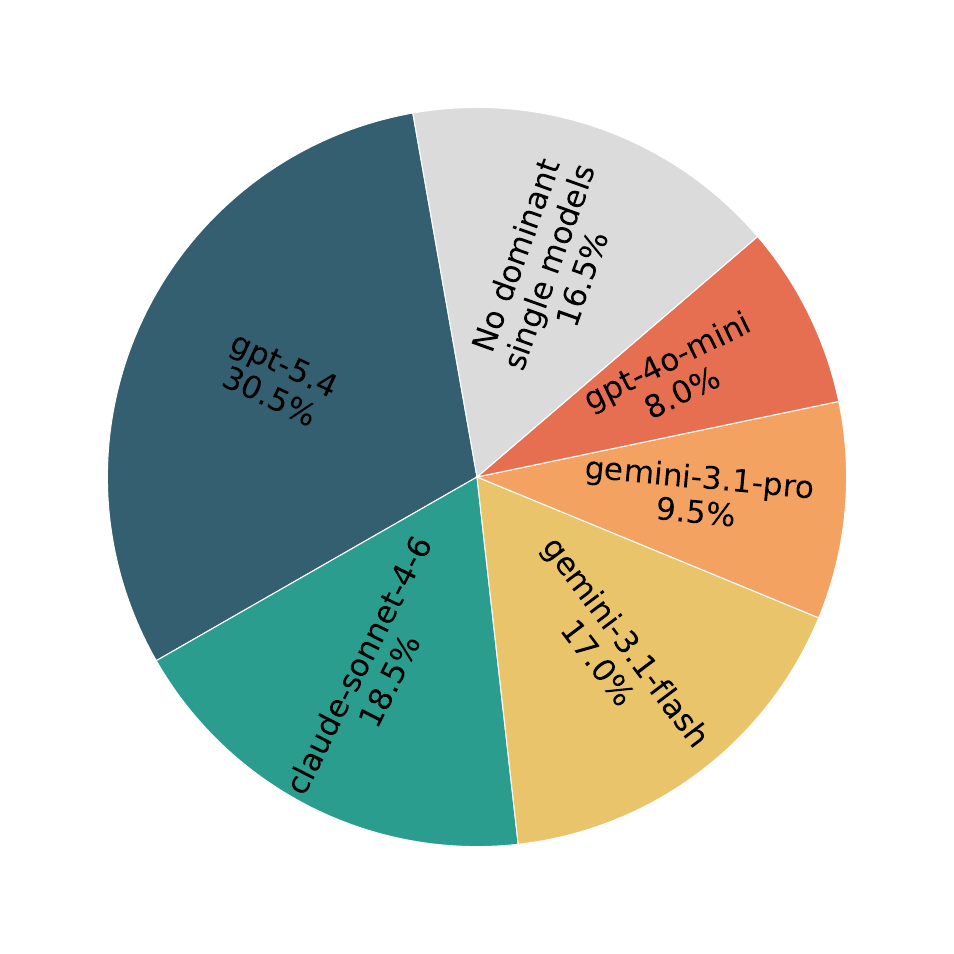}
    \caption{Frequency(\%) of each frontier LLM being the most diverse model. The threshold for deciding dominance is $5\%$. 
    }
    \label{fig:teaser_frontier}
\end{figure}

What does this dominance look like when using closed-source frontier LLMs?  We study five models on the test split of \wc: \texttt{claude-sonnet-4-6}, \texttt{gemini-3.1-flash-lite-preview}, \texttt{gemini-3.1-pro-preview}, \texttt{gpt-4o-mini} and \texttt{gpt-5.4}. We adhere to the same inference setups.  Figure \ref{fig:teaser_frontier} shows that even comparing among frontier models, we could not find an LLM that is dominantly more diverse than others for all queries. This further strengthens our argument that maximizing sample diversity requires ensembling from multiple models. However, looping in these SOTA models could be challenging for router training since it's difficult to probe hidden states. 

One might ask: how different are these frontier models from the set of open models in our routing pool? In Table \ref{tab:compare_frontier} we find that they have similar diversity coverage scores, lower uniqueness, higher answer length and quality. 
\begin{table*}[t]
    \footnotesize
    \centering
    {\setlength{\tabcolsep}{3.5pt}
    \begin{tabular}{lccccc}
    \toprule
        Model & Cov. & Qual & \#Unq & Avg \#Ans/Gen & Avg Ans. Len \\ \midrule
        \multicolumn{6}{c}{\textit{Closed Source}} \\ \midrule
        claude-sonnet-4-6             & 20.83\% & 5.0 & 24.2 & 7.2  & 220.4 \\
        gemini-3.1-flash-lite-preview & 21.48\% & 4.4 & 27.5 & 6.6  &  62.1 \\
        gemini-3.1-pro-preview        & 19.78\% & 4.4 & 23.7 & 6.7  & 162.1 \\
        gpt-4o-mini                   & 18.79\% & 3.9 & 28.1 & 8.7  &  44.4 \\
        gpt-5.4                       & 21.06\% & 4.7 & 27.0 & 7.4  & 262.2 \\ \midrule
        \multicolumn{6}{c}{\textit{Open Source}} \\ \midrule
        llama-3.2-1b                  &  9.72\% & 2.4 & 20.8 & 7.3  &  43.5 \\
        llama-3.2-3b                  & 10.17\% & 3.3 & 17.7 & 10.3 &  38.1 \\
        llama-3.1-8b                  & 11.52\% & 3.4 & 18.9 & 12.0 &  36.6 \\
        llama-3.3-70b                 & 12.02\% & 4.2 & 16.4 & 12.5 &  42.2 \\
        qwen2.5-72b-instruct          & 19.49\% & 4.5 & 25.2 & 5.8  &  83.8 \\
        qwen3-0.6b                    & 16.73\% & 3.0 & 31.4 & 4.4  &  30.1 \\
        qwen3-1.7b                    & 16.37\% & 3.9 & 25.5 & 8.3  &  46.3 \\
        qwen3-4b                      & 17.50\% & 4.2 & 25.0 & 10.7 &  50.4 \\
        qwen3-8b                      & 17.23\% & 4.0 & 26.0 & 11.5 &  38.0 \\
        qwen3-14b                     & 20.42\% & 4.1 & 29.7 & 14.0 &  47.6 \\
        gemma-3-1b-it                 & 18.74\% & 3.3 & 31.5 & 6.8  &  79.2 \\
        gemma-3-4b-it                 & 19.57\% & 3.5 & 31.9 & 11.7 &  56.9 \\
        gemma-3-12b-it                & 23.39\% & 3.9 & 33.9 & 11.7 &  88.0 \\
        gemma-3-27b-it                & 22.17\% & 4.3 & 29.2 & 10.4 & 123.9 \\
        olmo-2-0425-1b                & 23.83\% & 3.0 & 42.6 & 5.9  &  86.8 \\
        olmo-2-1124-7b                & 22.34\% & 3.7 & 33.8 & 7.5  &  53.4 \\
        olmo-2-1124-13b               & 21.82\% & 4.2 & 29.4 & 6.3  &  73.2 \\
        olmo-2-0325-32b               & 23.24\% & 4.3 & 30.4 & 5.8  &  74.7 \\
        \bottomrule
    \end{tabular}
    }
    \caption{Comparing 5 frontier models with 18 open models in our routing pool. Performance is calculated on test split of \wc.}
    \label{tab:compare_frontier}
\end{table*}

\section{Is there a dominant model on non-diversity questions?}

For the 18 models compared in the pilot study in Section \ref{sec:pilot}, we check their benchmark scores on general knowledge, math and reasoning questions, namely MMLU~\citep{hendrycks2020measuring} and GSM8K~\citep{cobbe2021gsm8k}. In addition to the standard accuracy scores, we also report the soft measure of log likelihood of the correct answers.  We implement the evaluation with \textbf{lm-evaluation-harness}~\citep{eval-harness}. 

Specifically, we follow the original evaluation setting of MMLU where each option's log likelihood is measured after the question prompt and use the most likely answer as model's prediction. For GSM8K, we conducted two forward passes. In the first pass, the model generates the reasoning process followed by the final answer at the end, by which we calculate the accuracy. In the second pass, we concatenate the question and the model's reasoning trajectory as prompt and measure the log likelihood of the ground truth answer as the output.   A model is considered to be dominant if its log likelihood of the correct answer is at least $5\%$ higher than the second best. 

The results are presented in Table \ref{tab:model_stats_mmlu} and Table \ref{tab:model_stats_gsm8k}. We don't find a dominant model on either benchmarks. But on MMLU we do see that using \texttt{Qwen2.5-72B-Instruct} dominates roughly $56\%$ questions.

\begin{table*}[t]
    \footnotesize
    \centering
    {\setlength{\tabcolsep}{3.5pt}
    \begin{tabular}{lcccc}
    \toprule
        Model & Mean LL & \%Best & \%Dom. & Raw Performance (Acc) \\ \midrule
        gemma-3-1b-it             & $-$2.553 &  2.14\% &  2.05\% & 38.66\% \\
        Qwen3-0.6B                & $-$2.553 &  0.28\% &  0.26\% & 40.34\% \\
        OLMo-2-0425-1B-Instruct   & $-$1.489 &  1.05\% &  0.93\% & 40.61\% \\
        Llama-3.2-1B-Instruct     & $-$1.245 &  0.80\% &  0.67\% & 48.40\% \\
        Qwen3-1.7B                & $-$2.089 &  1.67\% &  1.58\% & 55.44\% \\
        gemma-3-4b-it             & $-$2.495 &  4.70\% &  4.50\% & 57.57\% \\
        OLMo-2-1124-7B-Instruct   & $-$1.485 &  4.65\% &  4.37\% & 59.22\% \\
        Llama-3.2-3B-Instruct     & $-$0.957 &  0.68\% &  0.55\% & 62.18\% \\
        OLMo-2-1124-13B-Instruct  & $-$1.167 &  0.32\% &  0.30\% & 64.40\% \\
        Llama-3.1-8B-Instruct     & $-$0.811 &  0.46\% &  0.39\% & 68.31\% \\
        Qwen3-4B                  & $-$1.123 &  2.88\% &  2.61\% & 68.38\% \\
        gemma-3-12b-it            & $-$1.493 &  3.97\% &  3.70\% & 71.44\% \\
        Qwen3-8B                  & $-$1.081 &  6.88\% &  6.44\% & 73.02\% \\
        OLMo-2-0325-32B-Instruct  & $-$0.840 &  0.73\% &  0.68\% & 73.39\% \\
        gemma-3-27b-it            & $-$1.555 &  2.57\% &  2.44\% & 76.48\% \\
        Qwen3-14B                 & $-$0.810 &  5.46\% &  5.09\% & 77.10\% \\
        Llama-3.3-70B-Instruct    & $-$0.642 &  3.75\% &  3.50\% & 82.28\% \\
        Qwen2.5-72B-Instruct      & $-$0.933 & 57.00\% & 56.14\% & 83.35\% \\
        \bottomrule
    \end{tabular}
    }
    \caption{MMLU performance for the 18 open source models.\%Best denotes percentage of queries with highest log likelihood. \%Dom. denotes percentage of queries it dominates.}
    \label{tab:model_stats_mmlu}
\end{table*}

\begin{table*}[t]
    \footnotesize
    \centering
    {\setlength{\tabcolsep}{3.5pt}
    \begin{tabular}{lcccc}
    \toprule
        Model & Mean LL & \%Best & \%Dom. & Raw Performance (Acc) \\ \midrule
        Llama-3.2-1B-Instruct     & $-$9.752 &  0.15\% &  0.15\% & 16.38\% \\
        gemma-3-1b-it             & $-$11.766 &  0.15\% &  0.15\% & 24.49\% \\
        Qwen3-0.6B                & $-$8.773 &  0.15\% &  0.08\% & 32.83\% \\
        Llama-3.2-3B-Instruct     & $-$7.343 &  0.23\% &  0.23\% & 37.76\% \\
        Llama-3.1-8B-Instruct     & $-$7.059 &  0.08\% &  0.08\% & 47.76\% \\
        OLMo-2-0425-1B-Instruct   & $-$8.811 &  0.08\% &  0.08\% & 50.80\% \\
        Qwen3-1.7B                & $-$6.964 &  3.94\% &  3.87\% & 68.84\% \\
        gemma-3-4b-it             & $-$4.643 &  0.83\% &  0.83\% & 72.25\% \\
        OLMo-2-1124-7B-Instruct   & $-$5.577 &  0.61\% &  0.61\% & 73.01\% \\
        Llama-3.3-70B-Instruct    & $-$1.385 &  2.05\% &  1.82\% & 77.41\% \\
        OLMo-2-1124-13B-Instruct  & $-$4.801 &  1.14\% &  1.14\% & 79.53\% \\
        gemma-3-12b-it            & $-$3.203 &  1.14\% &  1.06\% & 79.61\% \\
        OLMo-2-0325-32B-Instruct  & $-$3.681 &  0.30\% &  0.30\% & 80.52\% \\
        Qwen3-4B                  & $-$2.599 &  1.21\% &  1.21\% & 86.96\% \\
        gemma-3-27b-it            & $-$2.699 & 20.39\% & 20.02\% & 87.26\% \\
        Qwen3-8B                  & $-$2.421 & 29.72\% & 28.81\% & 88.93\% \\
        Qwen3-14B                 & $-$1.403 &  5.69\% &  5.61\% & 92.49\% \\
        Qwen2.5-72B-Instruct      & $-$1.816 & 32.15\% & 30.17\% & 93.25\% \\
        \bottomrule
    \end{tabular}
    }
    \caption{GSM8K performance for the 18 open source models.}
    \label{tab:model_stats_gsm8k}
\end{table*}

\section{Prompts}
\label{app:prompts}

\begin{figure}[ht]
    \centering
    \vspace{-10pt}
    
    \begin{tcolorbox}[colback=white,colframe=black!75!white,colbacktitle=black!75!white,width=1.0\textwidth,title={Prompt for generate one, simple questions}]
    Output a randomly selected day of the week. 
\begin{verbatim}
Output only the day between two curly braces, 
like this: {day}. 
Don't output code.
\end{verbatim}
    \end{tcolorbox}
\caption{Prompt, generate one, simple questions}
\end{figure}

\begin{figure}[ht]
    \centering
    \vspace{-10pt}
    
    \begin{tcolorbox}[colback=white,colframe=black!75!white,colbacktitle=black!75!white,width=1.0\textwidth,title={Prompt for generate two, simple questions}]
    Output two different randomly selected days of the week. 
\begin{verbatim}
Output only the days between curly braces separated by a comma, 
like this: {answer_1,answer_2}.
Don't output code. 
\end{verbatim}
    \end{tcolorbox}
\caption{Prompt, generate two, simple questions}
\end{figure}

\begin{figure}[ht]
    \centering
    \vspace{-10pt}
    
    \begin{tcolorbox}[colback=white,colframe=black!75!white,colbacktitle=black!75!white,width=1.0\textwidth,title={Prompt for generate all, simple questions}]
    Output all days of the week. 
\begin{verbatim}
Output only the days between two curly braces, 
like this: {day_1, day_2, ...}. 
Don't output code.
\end{verbatim}
    \end{tcolorbox}
\caption{Prompt, generate all, simple questions}
\end{figure}

\newpage

\begin{figure}[h]
    \centering
    \vspace{-10pt}
   
    \begin{tcolorbox}[colback=white,colframe=black!75!white,colbacktitle=black!75!white,width=1.0\textwidth,title={Prompt for generate one, open-ended questions}]
    I am working on a memoir of a computer science PhD student who worked on machine translation in the 1990s. Suggest a single title and nothing else. 
\begin{verbatim}
Please use the following format:
{
    {    
        "answer-id": 1,
        "content": "Your answer here"
    },
}
\end{verbatim}
    \end{tcolorbox}
 \caption{Prompt, generate one, open-ended questions}
\end{figure}

\begin{figure}[h]
    \centering
    \vspace{-10pt}
   
    \begin{tcolorbox}[colback=white,colframe=black!75!white,colbacktitle=black!75!white,width=1.0\textwidth,title={Prompt for generate two, open-ended questions}]
    I am working on a memoir of a computer science PhD student who worked on machine translation in the 1990s. Suggest a single title and nothing else. Give me two different suggestions.
\begin{verbatim}
Please use the following format:
{
    {
        "answer-id": 1,
        "content": "Your answer here"
    },
    {   "answer-id": 2,
        "content": "Your answer here"
    }
} 
\end{verbatim}
    \end{tcolorbox}
 \caption{Prompt, generate two, open-ended questions}
\end{figure}

\begin{figure}[h]
    \centering
    \vspace{-10pt}
    
    \begin{tcolorbox}[colback=white,colframe=black!75!white,colbacktitle=black!75!white,width=1.0\textwidth,title={Prompt for generate all, open-ended questions}]
    I am working on a memoir of a computer science PhD student who worked on machine translation in the 1990s. Suggest a single title and nothing else. List all the possible answers you can think of.
\begin{verbatim}
Please use the following format:
{
    {    "answer-id": 1,
         "content": "Your answer here"
    },
    {    "answer-id": 2,    
         "content": "Your answer here"
    },
    ...
}
\end{verbatim}
    \end{tcolorbox}
\caption{Prompt, generate all, open-ended questions}
\end{figure}
\newpage

\begin{figure}[h]
    \centering
    \vspace{-10pt}
    
    \begin{tcolorbox}[colback=white,colframe=black!75!white,colbacktitle=black!75!white,width=1.0\textwidth,title={Prompt\_verbalized\_all, simple questions}]
    Output all days of the week.
\begin{verbatim}
For each output, also provide a numeric probability of sampling that output. 
Please sample at random from the full distribution.
Output only the days and probabilities between two curly braces, like this: 
{(day_1,probability_1), (day_2,probability_2) ...}. Don't output code."
\end{verbatim}
    \end{tcolorbox}
\caption{Prompt\_verbalized\_all, simple questions}
\end{figure}

\begin{figure}[h]
    \centering
    \vspace{-10pt}
   
    \begin{tcolorbox}[colback=white,colframe=black!75!white,colbacktitle=black!75!white,width=1.0\textwidth,title={Prompt\_verbalized\_all, open-ended questions}]
    I am working on a memoir of a computer science PhD student who worked on machine translation in the 1990s. 
    List all the possible answers you can think of. For each answer, also provide a numeric probability of sampling that answer.
\begin{verbatim}
Please use the following format:
{
    {
        \"answer-id\": 1,
        \"content\": \"Your answer here\",
        \"probability\": \"The probability of this answer\"
    },
    {
        \"answer-id\": 2,
        \"content\": \"Your answer here\", 
        \"probability\": \"The probability of this answer\"
    },
    ...
}
\end{verbatim}
    \end{tcolorbox}
 \caption{Prompt\_verbalized\_all, open-ended questions}
\end{figure}
\newpage

\begin{figure}[h]
    \centering
    \vspace{-10pt}
    
    \begin{tcolorbox}[colback=white,colframe=black!75!white,colbacktitle=black!75!white,width=1.0\textwidth,title={System prompt for system\_vanilla, simple questions and open-ended questions}]
    You are a helpful assistant. For each query, please generate all possible responses, each within a separate \texttt{\textless response\textgreater} tag. Responses should each include a \texttt{\textless text\textgreater}. 
    \end{tcolorbox}
\caption{System\_vanilla, simple questions and open-ended questions}
\end{figure}

\begin{figure}[h]
    \centering
    \vspace{-10pt}
   
    \begin{tcolorbox}[colback=white,colframe=black!75!white,colbacktitle=black!75!white,width=1.0\textwidth,title={System prompt for system\_verbalized\_all, simple questions and open-ended questions}]
    You are a helpful assistant. For each query, please generate all possible responses, each within a separate \texttt{\textless response\textgreater} tag. Responses should each include a \texttt{\textless text\textgreater} and a numeric \texttt{\textless probability\textgreater}. Please sample at random from the full distribution.
    \end{tcolorbox}
 \caption{System\_verbalized\_all, simple questions and open-ended questions}
\end{figure}



\section{Diversity coverage calculation details on open-ended questions}
\label{app:calc_details}
Here we discuss the details of how we evaluate the quality and diversity of answers to open-ended questions. We follow the exact procedure to partition the answer set and calculate. the quality scores in \citet{zhang2025noveltybench}. To determine semantic equivalence, we apply their equivalence classifier (used in line 5 in algorithm below)
to all pairs of generations and retain a subset with no mutually equivalent pairs (see Algorithm~\ref{alg:unique_generations} below). This classifier is finetuned with $1100$ pairs of  human annotated generations conditioned on prompts sampled from \nb and \wc.
We then score the quality of  each answer ${a'} \in {A_d}$, following their process: the score is first derived by a reward model and later mapped to $\{1, \ldots,10\} $\footnote{We use \texttt{Skywork-Reward-Gemma-2-27B-v0.2} model~\citep{liu2024skywork} as the reward model and the equivalence classifier released by~\citet{zhang2025noveltybench} at \url{https://huggingface.co/yimingzhang/deberta-v3-large-generation-similarity}.}. Their mapping is calibrated by aligning the distribution of reward model scores (from 2,400 MT-Bench generations) with GPT-4–judged quality scores, using thresholds to map reward values to the 1–10 scale. 

\begin{algorithm}[h]
\caption{Extract semantically nonequivalent generations}
\label{alg:unique_generations}
\begin{algorithmic}[1]
\Require Sampled generations $A = \{a_1, \dots, a_B\}$ 
\Require Equivalence classifier $\textsc{Eq}(\cdot,\cdot)$
\Ensure Set of semantically nonequivalent generations $A_d$

\State $A_d \gets \emptyset$

\For{$i = 1$ to $B$}
    \State $is\_duplicate \gets \textbf{False}$
    
    \For{each $a'\in A_d$}
        \State $s \gets \textsc{Eq}(a_i, a')$  \Comment{Similarity score}
        \If{$s > \tau$}
            \State $is\_duplicate \gets \textbf{True}$
            \State \textbf{break}
        \EndIf
    \EndFor
    
    \If{\textbf{not} $is\_duplicate$}
        \State $A_d \gets A_d \cup \{a_i\}$
    \EndIf
\EndFor

\State \Return $A_d$
\end{algorithmic}
\end{algorithm}

\section{Decoding settings}
\label{app:decoding}
We set target number ($N$) of answers  to $50$ if not otherwise stated. The temperature and top$-p$ are fixed to be $1.0$ and $1.0$ respectively. The max tokens is set to be $4096$. We use 2 H200 GPUs for all models. The batch size is $64$. We repeat the sampling process until $N$ answers are collected. The inference time varies by model sizes and families. We disable the thinking mode for \texttt{Qwen} models.

\section{Router implementation details}
\label{app:router_implementation}
We use Adam optimizer, a learning rate of $1e-3$.  For BERT classifier, we use the AdamW optimizer with a learning rate of $2e-5$. During training, we perform a grid search over options of \{soft, one-hot\} labels, weight decay and hidden dimensions.  Routers are selected based on the best scores on the validation set.  We experiment with soft labels and one-hot labels to provide the training signals. The \textbf{soft labels} are drawn by normalizing the diversity coverage scores against the most diverse model for this query. 
\textbf{One-hot labels} are  derived by $\mathbbm{1}[m_i = m_j^*]$.  We find that soft labels work best with M-way MLP classifier while one-hot labels are best for Binary MLP classifier.

\section{Scaling router training data}
\label{app:scaling_router}

Specifically, we experiment with training the router on $500$ and $1$K samples from \wc~, and $500$, $1$K, and $2$K samples from \ic~\citep{jiang2026artificial}. The results are shown in Table~\ref{tab:scaling_routing_data}. Increasing \wc~ training data from $500$ to $1$K improves diversity coverage on the \wc~test set, though it does not transfer to \ic. In contrast, scaling \ic~ data from $500$ to $2$K steadily improves performance on both the \ic~ test set and the \wc~ test set, indicating stronger generalization. Finally, jointly training on a combination of \wc~ and \ic~ further improves performance, slightly surpassing the best router ($26.4\%$ vs.\ $26.3\%$) trained on $1$K \wc~ data in Table~\ref{tab:router}.

\begin{table}[ht]
\small
    \centering
    \begin{tabular}{lllcc}
    \toprule
     &&& \multicolumn{2}{c}{\textbf{Evaluation Data}}\\
    \textbf{Method} &&& \wc  &   \ic \\ \midrule
         Random &&&18.13\% &18.24\%  \\
         Top Overall&& &23.83\% &23.13\%\\
         Oracle &&&33.04\% &30.50\% \\ \midrule
        &  \textbf{Training Data} & \textbf{Size} \\\midrule
    Router & \wc&500 &25.28\% $\scriptstyle \pm 0.28\% $ &22.58\% $\scriptstyle \pm 0.30\%$\\
       Router &   \wc &1K  &26.27\% $\scriptstyle \pm 0.13\%$ &22.58\% $\scriptstyle \pm 0.39\%$ \\ \midrule
      Router &    \ic&500 &23.98\% $\scriptstyle \pm 0.67\%$ &22.54\% $\scriptstyle \pm 0.60\%$\\ 
    Router &      \ic&1K &24.95\% $\scriptstyle \pm 0.28\%$ &23.54\% $\scriptstyle \pm 0.12\%$\\ 
      Router &    \ic&2K &25.13\% $\scriptstyle \pm 0.36\%$ &\textbf{23.78}\% $\scriptstyle \pm 0.16\%$\\  \midrule
        Router &  \wc~ and \ic &1K and 1K &26.05\% $\scriptstyle \pm 0.32\%$ &23.36\%  $\scriptstyle \pm 0.23\%$ \\
         Router & \wc~ and \ic & 1K and 2K &\textbf{26.40}\% $\scriptstyle \pm 0.21\%$ & 23.55\% $\scriptstyle \pm 0.10\%$ \\ \bottomrule

    \end{tabular}
    \caption{Router performance (diversity coverage) steadily improves with more training data.  We report the average and variance of 5 training runs with different random seeds. }
    \label{tab:scaling_routing_data}
\end{table}

\newpage

\section{Router Performance}

\begin{table*}[ht]
    \footnotesize
    \centering
    \begin{tabular}{lcccccccccc}
    \toprule
        \multirow{2}{*}{Method} & \multicolumn{5}{c}{NB-WildChat} & \multicolumn{5}{c}{NB-Curated (OOD)}  \\
        \cmidrule(lr){2-6} \cmidrule(lr){7-11}
         &Acc & \#U & Q & UQ & Cov. & Acc & \#U & Q & UQ & Cov.    \\ \midrule
         Top Overall &19.5\%  &42.6 &3.0 &2.9 &\textbf{23.8\%}  &3.4\% &35.4 &6.0 &5.7 &\textbf{38.6\%} \\
         Random M / Q &5.9\% &27.8 &3.7 &3.6 &18.1\% &5.6\% &27.8 &7.2 &7.0 &37.5\% \\
         Frequency &12.0\% &33.1 &3.8 &3.6 &21.0\%  &9.0\% &28.2 &7.2 &7.1 &39.6\%  \\
         
         Top M / Q  (oracle)  &100\% & 38.8 &4.5 &4.4 &33.0\%  & 100\% &30.3 &7.6 &7.4 &59.6\%  \\ \midrule
         $1$NN &16.5\% &34.3 &3.7 &3.6 &23.1\%  &5.6\% &28.2 &7.3 &7.1 &39.7\%   \\
         $5$NN &17.5\% &34.9 & 3.8 &3.7 &24.1\%  &12.4\% &29.8 &7.3 &7.1 &40.2\%  \\
         M-way BERT &22.0\% &40.3 &3.3 &3.2 &24.4\% &11.2\% &35.0 &6.3 &6.2 &40.3\%\\ 
         M-way MLP(agn) &24.0\% &35.1 &3.9 &3.8 &25.3\%  &12.4\% &30.1 &7.6 &7.5 &40.3\% \\
         M-way MLP(spec) & 27.0\% &39.3 &3.5 &3.4 &25.9\% &5.6\% &34.6 &6.3 &6.1 &40.2\% \\
         Binary MLP (agn) &23.9\% &38.4 &3.5 &3.4 &25.7\%$^{**}$ &10.8\% &32.8 &7.1 &7.0 &\textbf{40.7\%}$^{**}$ \\
         Binary MLP (spec) &23.9\% &38.1 &3.6 &3.5 &\textbf{26.3\%}$^{**}$ &13.3\% &30.8 &7.0 &6.8 &39.3\%$^{ns}$ \\ 
         \bottomrule
         
    \end{tabular}
     \caption[Caption]{A per-query router selecting over 18 models to maximize diversity coverage (Cov.).\#U denotes number of unique outputs, Q denotes average quality, and UQ denotes quality of unique outputs.  Accuracy(Acc) measures how frequently the router predicts the oracle model(ground truth target).  Random M / Q denotes random model per query, and Top M / Q denotes top model per query.}
     \label{tab:router_acc}
\end{table*}

\section{Distribution of Quality Scores for All and Unique Answers}
\label{app:quality_variance}

\begin{wrapfigure}{r}{0.48\linewidth}
    \centering
    \vspace{-2em}
    \includegraphics[width=\linewidth]{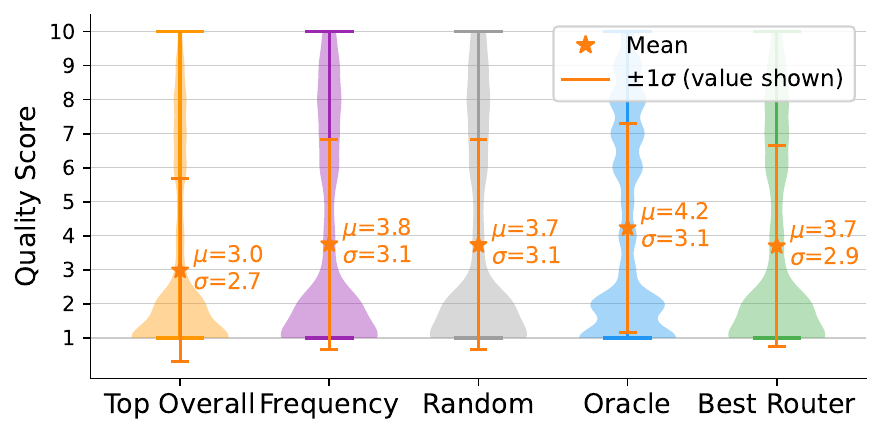}
    \vspace{-1.5em}
    \caption{Distribution of quality scores for all generated answers on \wc across different routing methods.}
    \label{fig:quality_dist_router}
    \vspace{-1em}
\end{wrapfigure}
\paragraph{Raw answer quality distribution}
Figure~\ref{fig:quality_dist_router} reports quality scores over all generated answers before deduplication. The trend is similar to the unique-answer quality distribution in Figure~\ref{fig:unique_quality_dist_router}: the learned router maintains competitive mean quality with relatively low variance among non-oracle methods.

\begin{figure}[h]
    \centering
    \includegraphics[width=0.95\linewidth]{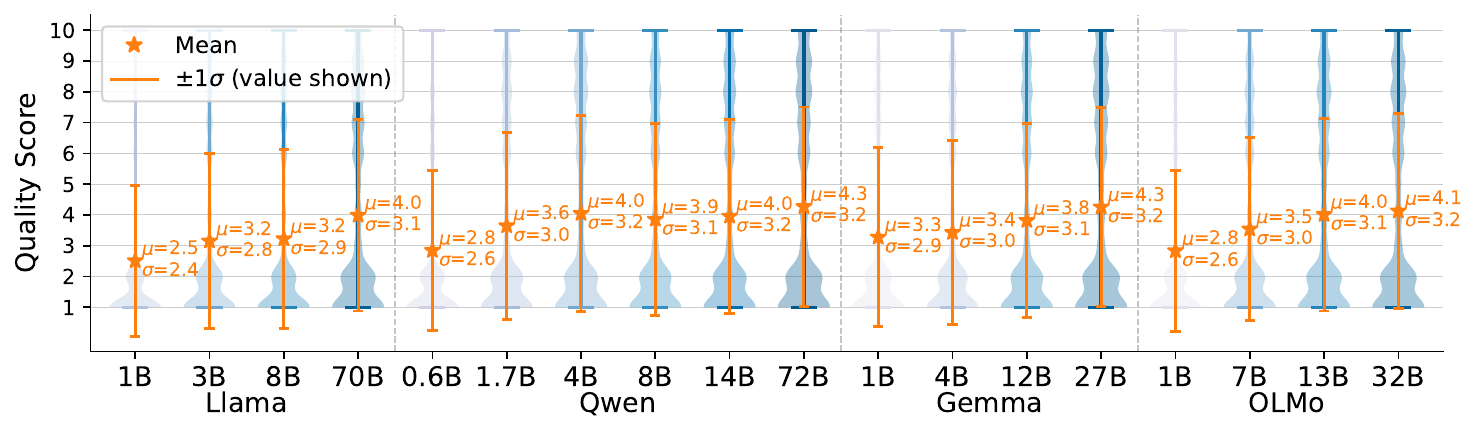}
    \caption{Distribution of raw answer quality scores across the 18 models on \wc.}
    \label{fig:quality_dist_model}
\end{figure}

\begin{figure}[h]
    \centering
    \includegraphics[width=0.95\linewidth]{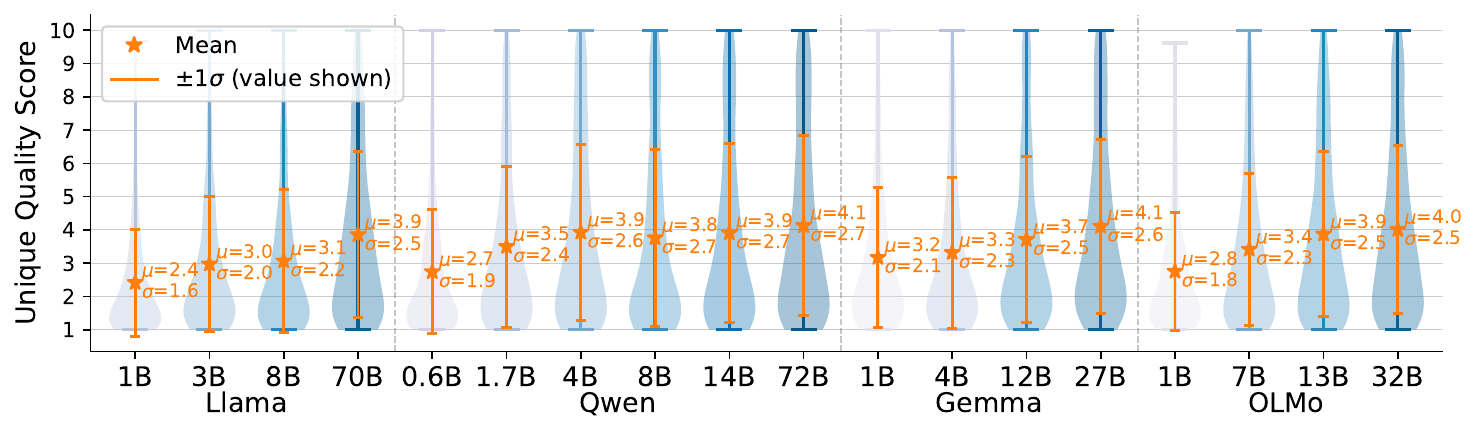}
    \caption{Distribution of unique-answer quality scores across the 18 models on \wc.}
    \label{fig:unique_quality_dist_model}
\end{figure}

\paragraph{Raw and unique answer quality by model.}
Figures~\ref{fig:quality_dist_model} and~\ref{fig:unique_quality_dist_model} compare the distributions of quality scores across the 18 individual models on \wc. For raw answers, larger models generally achieve higher mean quality, accompanied by greater variability. For example, Llama-70B reaches $\mu=4.0$ and $\sigma=3.1$, compared with $\mu=2.5$ and $\sigma=2.4$ for Llama-1B. This overall trend is visible across all four model families, although it is not strictly monotonic for every model size.

A similar pattern persists after deduplication. Within each model family, the mean quality and standard deviation of unique answers generally increase with model size. For example, Qwen-0.6B has $\mu=2.7$ and $\sigma=1.9$, whereas Qwen-72B reaches $\mu=4.1$ and $\sigma=2.7$. Similarly, OLMo-1B has $\mu=2.8$ and $\sigma=1.8$, compared with $\mu=4.0$ and $\sigma=2.5$ for OLMo-32B. These results suggest that larger models tend to produce higher-quality unique answers, but also a broader range of answer quality.

\section{Discussion}

\subsection{Different Prompt Templates}
\label{app:prompt_template}


Prompting methods affect generation diversity~\citep{zhang2025verbalized}.
We show that model ensembling is effective for answers generated by \textit{sequential} prompting: models are asked to generate as many distinct answers in one generation, where the latter answers are \textit{dependent} on previous answers. Does it also work for other prompting methods? We extend the study in section \ref{sec:pilot} to compare three different prompt types\footnote{Please refer to Appendix \ref{app:prompts} for the exact prompts.}:
\begin{itemize}[leftmargin=10px]
    \item \textbf{Generate one}: The model is prompted to produce \textit{one} random answer for the given question.
    \item \textbf{Generate two}: The model is prompted to provide two possible and different answers for given question.
    \item \textbf{Generate all (our default setting)}: The model is prompted to list out \textit{all} possible answers sequentially.  
\end{itemize}

\begin{table*}[ht]
 \footnotesize
     \centering
    \begin{tabular}{lccc ccc ccc}
    \toprule
    \multirow{2}{*}{Method} &\multicolumn{3}{c}{Gen 1} &\multicolumn{3}{c}{Gen 2} &\multicolumn{3}{c}{Gen All}\\
    \cmidrule(lr){2-4} \cmidrule(lr){5-7} \cmidrule(lr){8-10} 
    &\#Unq &Quality  &Cov. &\#Unq &Quality  &Cov. &\#Unq &Quality &Cov. \\ \midrule
    Random &13.7 &4.9 &9.9\% &18.3 &4.5 &13.2\% &27.8 &3.7 &18.1\% \\
    Frequency &25.4 &3.9 &15.6\% &25.2 &4.1 &17.1\% &33.1 &3.8 &21.0\% \\
    Top Overall &31.8 &3.2 &18.5\% &34.8 &3.1 &19.7\% &42.6 &3.0 &23.8\%\\ \midrule
    Oracle (G-1) & \cellcolor{LightGrey} 32.8 &\cellcolor{LightGrey}4.1 &\cellcolor{LightGrey}25.6\% &\cellcolor{LightGrey!40}32.1 &\cellcolor{LightGrey!40}3.7 &\cellcolor{LightGrey!40}22.3\% &\cellcolor{LightGrey!40}31.0 &\cellcolor{LightGrey!40}3.3 &\cellcolor{LightGrey!40}20.4\%\\
    Oracle (G-2) &\cellcolor{LightGrey!40}24.1 &\cellcolor{LightGrey!40}4.3 &\cellcolor{LightGrey!40}19.5\% &\cellcolor{LightGrey}32.3 &\cellcolor{LightGrey}4.7 &\cellcolor{LightGrey}28.3\% &\cellcolor{LightGrey!40}31.6 &\cellcolor{LightGrey!40}3.3 &\cellcolor{LightGrey!40}21.0\% \\
    Oracle (G-All)  &\cellcolor{LightGrey!40}16.4 &\cellcolor{LightGrey!40}5.0 &\cellcolor{LightGrey!40}14.7\% &\cellcolor{LightGrey!40}23.1 &\cellcolor{LightGrey!40}4.6 &\cellcolor{LightGrey!40}18.8\% &\cellcolor{LightGrey}38.8 &\cellcolor{LightGrey}4.5 &\cellcolor{LightGrey}33.0\%  \\
    \midrule
    Router (G-1) &\cellcolor{LightGrey}33.1 &\cellcolor{LightGrey}3.2 &\cellcolor{LightGrey}19.1\% &\cellcolor{LightGrey!40}33.3 &\cellcolor{LightGrey!40}3.3 &\cellcolor{LightGrey!40}20.4\% &\cellcolor{LightGrey!40}25.8 &\cellcolor{LightGrey!40}2.8 &\cellcolor{LightGrey!40}14.4\% \\ 
    Router (G-2) &\cellcolor{LightGrey!40}26.5 \cellcolor{LightGrey!40}&\cellcolor{LightGrey!40}4.2 &\cellcolor{LightGrey!40}\textbf{19.7\%} &\cellcolor{LightGrey}30.5 &\cellcolor{LightGrey}3.9 &\cellcolor{LightGrey}\textbf{21.6\%} &\cellcolor{LightGrey!40}29.2 &\cellcolor{LightGrey!40}3.2 &\cellcolor{LightGrey!40}19.0\%\\
    \makecell[l]{Router (G-All)} &\cellcolor{LightGrey!40}18.3 &\cellcolor{LightGrey!40}4.7 &\cellcolor{LightGrey!40}14.8\% &\cellcolor{LightGrey!40}24.4 &\cellcolor{LightGrey!40}4.3 &\cellcolor{LightGrey!40}18.1\% &\cellcolor{LightGrey}37.5 &\cellcolor{LightGrey}3.7 &\cellcolor{LightGrey}\textbf{26.2\%} \\
    \bottomrule
    \end{tabular}
    \caption[Caption]{Training the router under different prompting strategies (\colorbox{LightGrey}{in domain} and \colorbox{LightGrey!40}{out-of-domain}  evaluation) on \wc. Router (X) is a router trained under prompt type X. Oracle (X) denotes that we always use ground truth labels of prompt X  as predictions.
    Training a router improves diversity for all prompts, as all routers beat their Top Overall baselines. However, different prompt templates seem to elicit different levels of diversity in LLMs, as the oracle predictions don't generalize across prompts. 
     }
     \label{tab:router_prompt_full}
\end{table*}

\paragraph{Routing improves diversity for all prompts, yet a router trained on one prompt does not generalize to others.} We ablate the prompting strategies, retrain routers, and evaluate them on all types of prompts. The performance is presented in  Table \ref{tab:router_prompt_full}. We find that generate all prompt incurs most diversity coverage, as shown by the diversity scores (Cov.) of the random/ oracle baselines. Training a router in-domain consistently improves diversity coverage, yet neither oracle labels nor routers generalize across prompts. Finally, larger gains are observed when routing under better prompts.


\begin{table*}[h]
    \footnotesize

    \vskip 0.15in
    \centering
    \begin{tabular}{l ccc ccc }
         \toprule
          &\multicolumn{2}{c}{Gen 1} &\multicolumn{2}{c}{Gen 2} &\multicolumn{2}{c}{Gen All}\\ 
           &Cov. &Len  &Cov. &Len &Cov. &Len \\
         \midrule
          Random &$16.7\%$ &$49.3$  &$22.4\%$ &$37.0$  &$37.5\%$ &$17.1$ \\
          Top model &$33.5\%$ &$47.7$  &$35.4\%$ &$38.8$ &$47.0\%$ &$22.6$  \\
          Top 2 models &$32.7\%$ &$31.6$  &$36.1\%$ &$34.2$  &$45.6\%$ &$24.0$ \\
          Top model per query  &$43.6\%$ &$37.8$  &$46.6\%$ & $30.3$ &$59.6\%$ &$21.9$ \\
         \bottomrule
    \end{tabular}
    \caption[Caption]{Oracle diversity coverage (Cov.) and answer lengths (Len) for different prompt templates on \nb Questions.
    }
    \label{tab:ensemble_length}
\end{table*}

\begin{figure}[h]
    \centering
\includegraphics[width=0.8\linewidth]{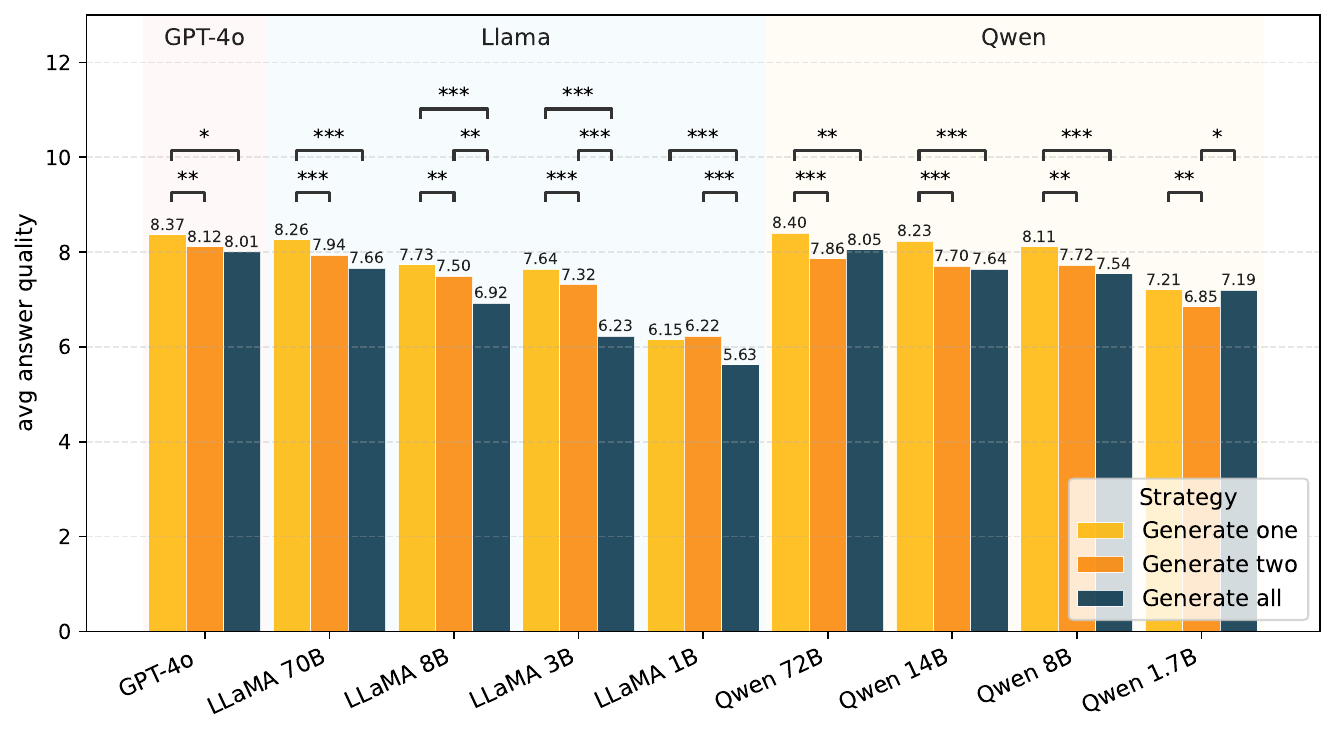}
    \vspace{-1em}
    \caption{Average answer quality for responses generated from different prompts on \nb questions.}
    \label{fig:tradeoff}
\end{figure}

\paragraph{Tradeoff of the generate all prompt.}
Despite being the best method, there is a trade-off between diversity and quality for generate all.  Under the routing setting,  we observe in Table~\ref{tab:ensemble_length} that the length of the answers decreases from generate one, generate two to generate all, up to $66\%$ (from $49.3$ to $17.1$).   Besides, as shown in Table \ref{tab:router_prompt_full}, though the number of unique answers sampled increases, the average answer quality deteriorates from generate one, generate two to generate all.
This claim is further supported
by comparing average answer quality among different prompting methods in Figure~ \ref{fig:tradeoff}. It shows that generate all has the lowest answer quality while generate one has the highest. These findings hold for models across different sizes and families.  Interestingly, a closer look into the answer generation process suggests that answers generated at later positions have worse quality under the sequential generate all prompt in Figure \ref{fig:score_position}.

\subsection{Discussions: Other configurations/hyperparameters that we can vary}
\label{app:configs}

\begin{table}[h]
    \centering
    \begin{tabular}{lcccc}
        \toprule
        Ratio & \#Unq & Qual & Unq Qual & Cov. \\
        \midrule
        \multicolumn{5}{c}{\textit{Oracle model pair}} \\
         0:50  & 42.50 & 4.16 & 3.95 & 35.40\% \\
         5:45  & 43.90 & 4.11 & 3.94 & 36.82\% \\
         10:40 & 45.00 & 4.13 & 3.99 & 37.56\% \\
         15:35 & 45.50 & 4.09 & 3.88 & 37.10\% \\
         20:30 & 46.30 & 3.93 & 3.80 & 36.46\% \\
         25:25 & 45.30 & 4.00 & 3.87 & 36.40\% \\
        \midrule
        \multicolumn{5}{c}{\textit{Top 2 model pair}} \\
         0:50  & 36.70 & 3.74 & 3.53 & 24.36\% \\
         5:45  & 38.10 & 3.73 & 3.46 & 25.58\% \\
         10:40 & 39.20 & 3.58 & 3.34 & 25.60\% \\
         15:35 & 40.50 & 3.40 & 3.20 & 25.34\% \\
         20:30 & 42.10 & 3.33 & 3.15 & 26.28\% \\
         25:25 & 43.60 & 3.26 & 3.12 & 26.98\% \\
         30:20 & 44.30 & 3.12 & 3.04 & 26.54\% \\
         35:15 & 45.60 & 3.06 & 3.02 & 27.48\% \\
         40:10 & 46.20 & 2.99 & 2.98 & 27.62\% \\
         45:5  & 46.70 & 2.91 & 2.90 & 27.10\% \\
         50:0  & 47.20 & 2.85 & 2.88 & 27.10\% \\
         \textit{avg} & 42.75 & 3.27 & 3.15 & 26.36\% \\
        \midrule
        \multicolumn{5}{c}{\textit{Random model pair}} \\
         0:50  & 35.93 & 3.24 & 3.23 & 20.86\% \\
         5:45  & 36.56 & 3.26 & 3.24 & 21.51\% \\
         10:40 & 37.19 & 3.23 & 3.21 & 21.72\% \\
         15:35 & 37.49 & 3.22 & 3.20 & 21.79\% \\
         20:30 & 37.73 & 3.23 & 3.20 & 21.86\% \\
         25:25 & 37.72 & 3.22 & 3.19 & 21.89\% \\
        \bottomrule
    \end{tabular}
    \caption{Exploring different ratios while varying models per question. Performance is reported on 10 questions sampled from \wc. Top 2 models are \texttt{olmo-2-0425-1b} and \texttt{olmo-2-0325-32b}.}
    \label{tab:fix_ratio_vary_model}
\end{table}

\begin{table}[h]
    \centering
    \begin{tabular}{lcccc}
        \toprule
        Strategy & \#Unq & Qual & Unq Qual & Cov. \\
        \midrule
        Oracle ratio          & 44.80 & 3.66 & 3.55 & 32.58\% \\
        Overall best (40:10)  & 46.20 & 2.99 & 2.98 & 27.62\% \\
        Half/half             & 43.60 & 3.26 & 3.12 & 26.98\% \\
        \bottomrule
    \end{tabular}
    \caption{Always use top 2 models ( \texttt{olmo-2-0425-1b} and \texttt{olmo-2-0325-32b}), while varying ratios per question.}
    \label{tab:fix_model_vary_ratio}
\end{table}

\paragraph{More flexible proportions of sampling per model} In the previous setting of routing to two models, we fix the sampled answers to be split equally (i.e., if there are two models selected to generate 50 answers, each would contribute to 25 answers).  Will a more flexible proportion lead to more diversity?
 Under the same setting of sampling 50 answers from two models, we experiment with a set of possible ratios 0.0:1.0, 0.1:0.9, 0.2:0.8, 0.3:0.7, 0.4:0.6, 0.5:0.5(original) to assign the budget between two models. We conduct two experiments: (1) pick a ratio for all the questions, vary model choices (2) fix two models to ensemble (top 2 by individual performance), varying ratios for each question. We present the results in table \ref{tab:fix_ratio_vary_model} and table \ref{tab:fix_model_vary_ratio} respectively. We find that for oracle/random/top 2 model pairs, different global ratios don't have much difference in output diversity. If we fix 2 models to ensemble and optimize ratios for each question, the score can be improved over rigid half/half mixing ($32.58\%$ vs $26.98\%$).

\begin{table}[h]
    \centering
    \begin{tabular}{lcccc}
        \toprule
        $N$ & \#Unq & Qual & Unq Qual & Cov. \\
        \midrule
        1  & 42.50 & 4.16 & 3.95 & 35.40\% \\
        2  & 45.30 & 4.00 & 3.87 & 36.08\% \\
        3  & 44.00 & 4.04 & 3.86 & 35.72\% \\
        4  & 44.20 & 4.13 & 3.79 & 35.96\% \\
        5  & 43.80 & 4.13 & 3.79 & 35.74\% \\
        6  & 44.30 & 3.93 & 3.58 & 35.04\% \\
        7  & 43.40 & 4.08 & 3.76 & 35.12\% \\
        8  & 43.30 & 3.94 & 3.70 & 34.02\% \\
        9  & 42.90 & 4.02 & 3.73 & 34.00\% \\
        10 & 43.10 & 3.92 & 3.65 & 33.36\% \\
        11 & 42.40 & 3.87 & 3.66 & 31.82\% \\
        12 & 42.40 & 3.80 & 3.56 & 30.92\% \\
        13 & 40.40 & 3.83 & 3.65 & 29.56\% \\
        14 & 40.10 & 3.80 & 3.63 & 28.88\% \\
        15 & 38.80 & 3.70 & 3.58 & 27.38\% \\
        16 & 37.90 & 3.66 & 3.53 & 26.10\% \\
        17 & 37.30 & 3.73 & 3.54 & 25.66\% \\
        18 & 36.60 & 3.67 & 3.56 & 24.32\% \\
        \bottomrule
    \end{tabular}
    \caption{Oracle models: fix $N$ models to select for all questions and vary model choices per question.}
    \label{tab:n_models_oracle}
\end{table}

\begin{table}[h]
    \centering
    \begin{tabular}{lcccc}
        \toprule
        $N$ & \#Unq & Qual & Unq Qual & Cov. \\
        \midrule
        1  & 47.20 & 2.85 & 2.88 & 27.10\% \\
        2  & 43.60 & 3.26 & 3.12 & 26.98\% \\
        3  & 41.80 & 3.21 & 3.06 & 24.90\% \\
        4  & 41.90 & 3.31 & 3.07 & 25.40\% \\
        5  & 40.60 & 3.47 & 3.27 & 25.42\% \\
        6  & 40.70 & 3.53 & 3.34 & 26.60\% \\
        7  & 40.90 & 3.59 & 3.42 & 27.00\% \\
        8  & 40.00 & 3.65 & 3.54 & 27.26\% \\
        9  & 38.40 & 3.85 & 3.60 & 26.48\% \\
        10 & 38.90 & 3.84 & 3.64 & 27.20\% \\
        11 & 37.70 & 3.81 & 3.52 & 25.70\% \\
        12 & 38.50 & 3.62 & 3.30 & 25.00\% \\
        13 & 35.20 & 3.70 & 3.48 & 23.22\% \\
        14 & 37.20 & 3.57 & 3.35 & 23.96\% \\
        15 & 36.40 & 3.60 & 3.44 & 23.98\% \\
        16 & 35.80 & 3.67 & 3.48 & 23.82\% \\
        17 & 36.40 & 3.38 & 3.28 & 21.98\% \\
        18 & 34.90 & 3.57 & 3.52 & 22.50\% \\
        \bottomrule
    \end{tabular}
    \caption{Top $N$ models: fix $N$ models to select for all questions and vary model choices per question.}
    \label{tab:n_models_top}
\end{table}

\begin{table}[h]
    \centering
    \begin{tabular}{lcccc}
        \toprule
        $N$ & \#Unq & Qual & Unq Qual & Cov. \\
        \midrule
        1  & 35.93 & 3.24 & 3.23 & 20.86\% \\
        2  & 37.72 & 3.22 & 3.19 & 21.85\% \\
        3  & 38.63 & 3.23 & 3.17 & 22.38\% \\
        4  & 39.03 & 3.21 & 3.13 & 22.60\% \\
        5  & 39.11 & 3.22 & 3.13 & 22.89\% \\
        6  & 38.89 & 3.29 & 3.18 & 23.36\% \\
        7  & 38.53 & 3.34 & 3.22 & 23.58\% \\
        8  & 38.33 & 3.37 & 3.26 & 23.70\% \\
        9  & 38.35 & 3.44 & 3.32 & 24.08\% \\
        10 & 38.05 & 3.47 & 3.34 & 24.26\% \\
        11 & 37.82 & 3.46 & 3.32 & 23.66\% \\
        12 & 37.45 & 3.49 & 3.36 & 23.80\% \\
        13 & 37.44 & 3.53 & 3.39 & 23.70\% \\
        14 & 36.93 & 3.54 & 3.41 & 23.49\% \\
        15 & 36.46 & 3.58 & 3.44 & 23.49\% \\
        16 & 35.94 & 3.59 & 3.45 & 23.32\% \\
        17 & 36.66 & 3.63 & 3.51 & 23.92\% \\
        18 & 36.60 & 3.67 & 3.56 & 24.32\% \\
        \bottomrule
    \end{tabular}
    \caption{Random models: fix $N$ models to select for all questions and vary model choices per question.}
    \label{tab:n_models_random}
\end{table}

\begin{table}[h]
    \centering
    \begin{tabular}{lcccc}
        \toprule
        Strategy & \#Unq & Qual & Unq Qual & Cov. \\
        \midrule
        Oracle $N$              & 44.10 & 3.93 & 3.73 & 34.24\% \\
        Best overall ($N=8$)    & 40.00 & 3.65 & 3.54 & 27.26\% \\
        Random $N$              & 39.23 & 3.53 & 3.35 & 25.25\% \\
        \bottomrule
    \end{tabular}
    \caption{Fix model order to ensemble (ranking of individual performance) and vary $N$ per question.}
    \label{tab:n_models_vary_n}
\end{table}
\newpage

\paragraph{Varying the number of models to ensemble from}
In previous experiments, we fix the number of activated models to be $1$ (routing the best model per query) or $2$ (routing to two best models per query). Will sampling answers from more models, while keeping the total number of answers unchanged, improve diversity? We answer this question by two experiments: (1) fix the number of models $N$ for all, vary the selected models per question; (2) fix the order of model to be selected (ranked by individual performance), vary the number $N$ per question. We present the results in Table \ref{tab:n_models_oracle}, Table \ref{tab:n_models_top}, Table \ref{tab:n_models_random}, and Table \ref{tab:n_models_vary_n}.  We find that routing to a custom model per question remains the most promising approach (under oracle settings). Routing to two models can offer further gains. But ensembling $>$ 2 models does not improve output diversity.
\newpage

\paragraph{Scaling the number of candidates and generations}
In this paper, we study selecting models from a pool of 18 candidates. However, in a real-world setting, there are hundreds of models users can choose from. Therefore, future work can explore employing a larger pool of LLMs that better harness their complementary strengths of uncovering more diverse answers. Besides, the number of answers to open-ended questions is infinitely large, and future work is encouraged to explore sample sizes beyond $50$.

\section{Extended related work}
\paragraph{Measuring output diversity}  
Traditional metrics to measure lexical diversity and text style are based on token and POS n-grams statistics \citep{roemmele2017evaluating, see-etal-2019-massively, tevet-berant-2021-evaluating, meister2023locally} and embedding similarity between candidates~\citep{padmakumar2023does}. Later works go beyond the distinctness of outputs and also measure the validity of each response. 
\citet{zhang2024forcing} propose to evaluate the diversity of LLMs by calculating the coverage of gold targets and the KL-divergence from the desired distribution. However, providing ground-truth distributions for open-ended questions is non-trivial. 
Closely related to our work, \citet{zhang2025noveltybench} introduce the notion of \emph{user-perceived utility}, which jointly models uniqueness and quality while accounting for user patience. 
In this framework, uniqueness is computed by partitioning sampled answers into non-equivalent groups, and answer quality is estimated using reward model scores. 
However, this metric penalizes later-generated responses, whereas our goal is to assess how well a set of answers covers the answer space regardless of generation order.

Similarly, \citet{sorensen2025spectrum} evaluates how well a model covers an open-ended output space using validity and diversity metrics. 
However, their evaluation relies on expensive human annotations and thus is only experimented with a single model with four generations per prompt. 
In this work, we build on the framework of \citet{zhang2025noveltybench} and propose \textit{diversity coverage}, a metric that evaluates how well a set of generated answers covers the valid answer space across many generations and multiple LLMs without requiring additional human supervision.




\newpage
\section{Verbalized Sampling}
\label{app:vs}
Similar to our baselines, Verbalized Sampling is a recent prompting technique that increases LLM output diversity.  We decided not to include it in the main experiments since it performs similarly (if not worse than) our generated all baseline. We include the evidence  below in Table \ref{tab:vs_simple} and Table \ref{tab:vs_nb}:

\begin{table}[ht]
    \centering
   
    \label{tab:vs_simple}
    \begin{tabular}{lccccccc}
        \toprule
         \multirow{2}{*}{\textsc{Prompt}}&  \multirow{2}{*}{\textsc{Model}}& 
         \multicolumn{6}{c}{Cov. \%} \\
         \cmidrule(lr){3-8}
          & &1 &10 &20 &50 &100 &1000 \\ \midrule
         prompt\_vanilla & \multirow{4}{*}{Llama 8B} &$6.09$ &$47.96$ &$66.70$ &$88.56$ &$92.44$ &$96.51$ \\
         prompt\_verbalized\_all & &$6.09$ &$47.96$ &$66.69$ &$89.83$ &$93.81$ &$97.76$   \\
         system\_vanilla & &$6.09$ &$43.42$ &$55.31$ &$72.00$ &$83.02$ &$95.23$ \\
         system\_verbalized\_all & &$6.09$ &$44.11$ &$60.26$ &$79.16$ &$89.30$ &$94.98$\\ \midrule
          prompt\_vanilla &    \multirow{4}{*}{GPT-4o} &$6.09$ &$48.20$ &$67.18$ &$89.80$ &$92.80$ &$94.10$\\
         prompt\_verbalized\_all & &$6.09$ &$46.60$ &$62.27$ &$84.26$ &$87.15$ &$92.21$ \\
         system\_vanilla & &$6.09$ &$43.83$ &$54.77$ &$68.12$ &$76.33$ &$91.21$ \\
         system\_verbalized\_all & &$6.09$ &$44.46$ &$56.07$ &$68.59$ &$77.62$ &$92.00$ \\ 
         \bottomrule
    \end{tabular}
     \caption{Compared results with verbalized sampling on \sq generating  up to $1,10,20,...1000$ answers. Prompt\_vanilla is the existing generate-all prompt. System\_verbalized\_all is the original prompt proposed in verbalized sampling.}
\end{table}

\begin{table}[h]
    \centering
    \label{tab:vs_nb}
    \begin{tabular}{lcccccccc}
        \toprule
         \multirow{2}{*}{\textsc{Prompt}}&  \multirow{2}{*}{\textsc{Model}}& 
         \multicolumn{7}{c}{Cov. \%} \\
         \cmidrule(lr){3-9}
          & &1 &5 &10 &20 &50 &100 &200\\ \midrule
         prompt\_vanilla &  \multirow{4}{*}{Llama 8B} & $0.38$ &$1.67$ &$3.04$ &$5.22$ &$10.00$ &$16.17$ &$26.13$ \\
         prompt\_verbalized\_all & &$0.37$ &$1.63$ &$2.98$ &$4.99$ &$9.24$ &$14.46$ &$23.76$ \\
         system\_vanilla & &$0.34$ &$1.34$ &$2.42$ &$4.19$ &$8.91$ &$14.86$ &$24.55$ \\
         system\_verbalized\_all & &$0.37$ &$1.57$ &$2.82$ &$4.54$ &$9.05$ &$15.31$ &$25.53$ \\ \midrule
         prompt\_vanilla & \multirow{4}{*}{GPT-4o}  &$0.42$ &$1.84$ &$3.51$ &$6.30$ &$11.01$ &$16.47$ &$24.78$\\
         prompt\_verbalized\_all & &$0.40$ &$1.86$ &$3.29$ &$5.00$ &$8.65$ &$13.09$ &$19.61$ \\
         system\_vanilla & &$0.41$ &$1.57$ &$2.69$ &$4.08$ &$7.30$ &$11.05$ &$16.94$ \\
         system\_verbalized\_all & &$0.40$ &$1.71$ &$2.68$ &$4.16$ &$7.30$ &$11.12$ &$17.50$ \\ 
         \bottomrule
    \end{tabular}
    \caption{Compared results with verbalized sampling on \nb generating up to $1,5,10,...200$ responses.Prompt\_vanilla is the existing generate-all prompt. System\_verbalized\_all is the original prompt proposed in verbalized sampling.}
\end{table}

\clearpage
\section{Generating diverse outputs out of a single model}

    
    

\begin{figure*}[h]
    \centering
    \includegraphics[width=0.9\linewidth]
    {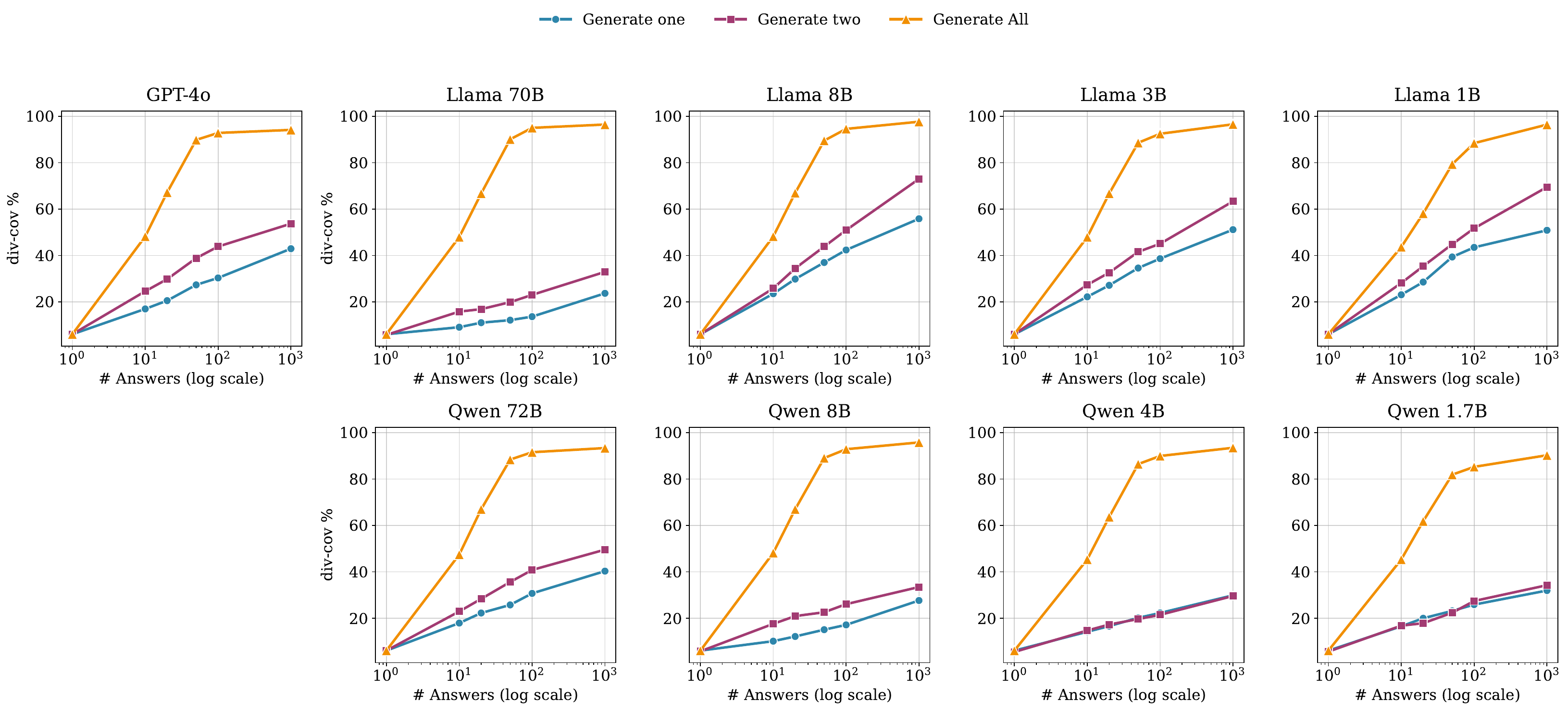}
    
    \includegraphics[trim=0 0 0 1cm,clip,width=0.9\linewidth]{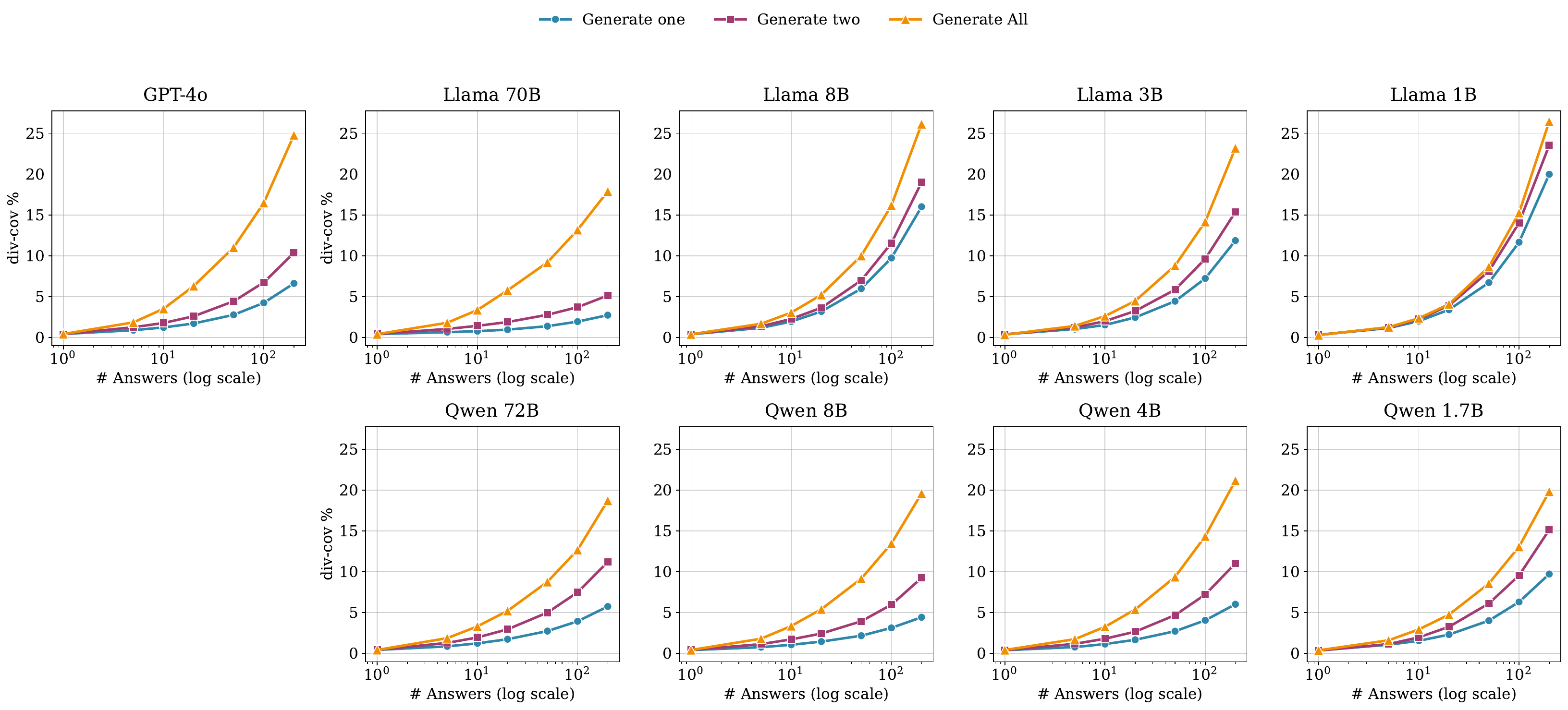}
    
    \caption{Comparing different prompts for generating diverse outputs. The number of samples (x-axis) vs. the diversity coverage scores (y-axis). 
    We compare three prompts on \sq (top) and \nb (bottom) datasets.}
    \label{fig:scaling} 
\end{figure*}

\begin{figure*}[h]
    \centering
    \includegraphics[width=0.9\linewidth]{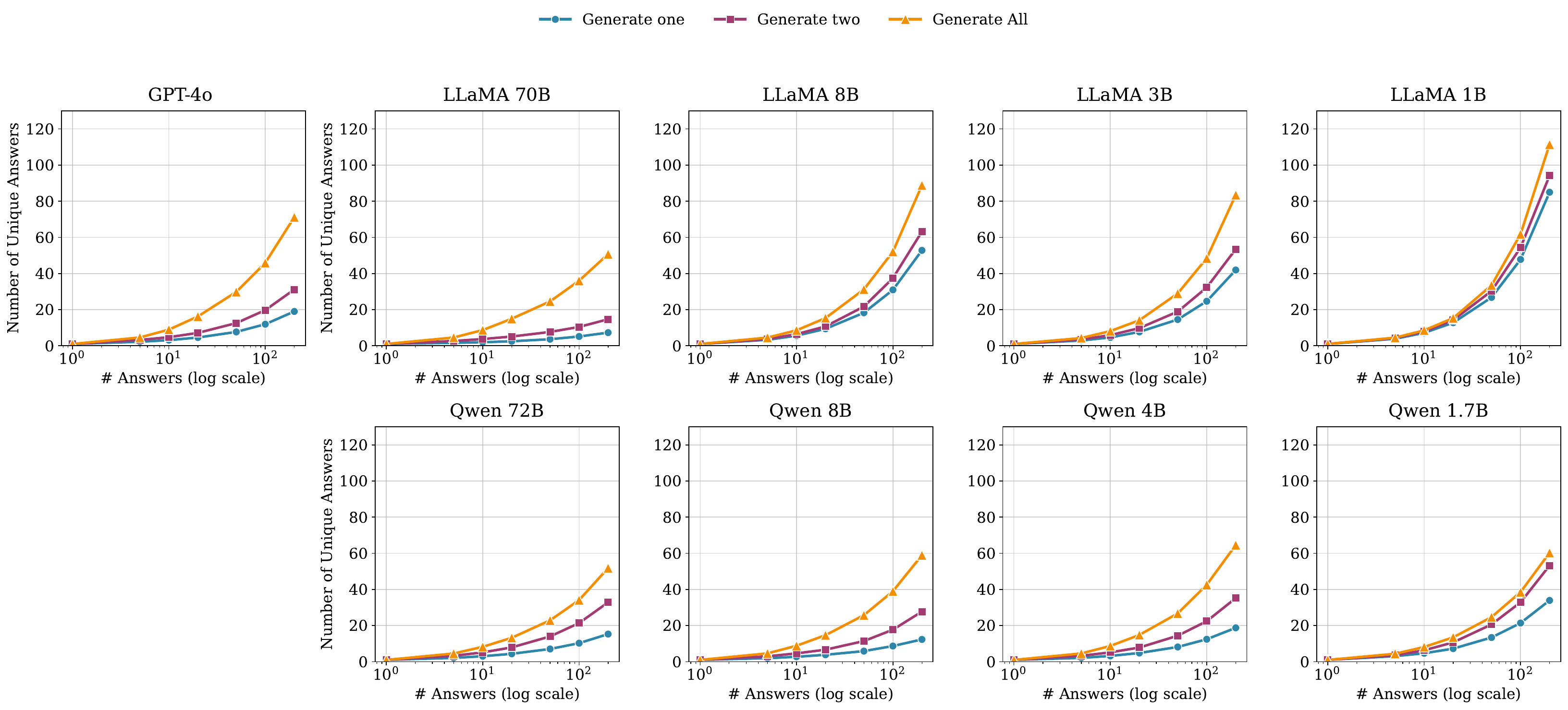}
    \caption{The number of unique answers on \nb. The X-axis is the log of the number of answers generated. Y-axis measures the diversity coverage of all the unique answers divided by the max possible score(generate 200 different good answers).  
    }
    \label{fig:unique}
\end{figure*}
\newpage


\subsection{Experiment settings}

\paragraph{Decoding settings} For each prompting strategy and desired number of answers $N$,  we repeatedly sample generations from the model until we collect $N$ answers. For \sq, we use $N=\{1, 10, 20, 50, 100, 1000\}$. For \nb, we use $N=\{1, 5, 10,20, 50, 100, 200\}$.  We set the temperature to $1.0$,  top\_p to $1.0$. The  max\_len is set to 2048 by default. For the generate all setting in \nb, we extend the max\_len to 4096 because generations can not  be finished within 2048 tokens.

\subsection{Results}

\paragraph{Compare different prompting strategies}
Figure \ref{fig:scaling} shows how different prompting strategies  affect the diversity of combined answers.
For all models on both datasets, sequential generation enables a lot more answer diversity than parallel methods. With the best prompt, models on \sq saturate to more than $90\%$ of coverage rate. This suggests that for easy diversity questions, nearly all models have good knowledge of the full answer space. On \nb, answer diversity keeps growing as more generations are inferred. This reveals large diversity potential in uncovering more unique and high-quality responses to open-ended queries.

\begin{figure}[h]
    \centering
    
    \includegraphics[width=0.8\linewidth]{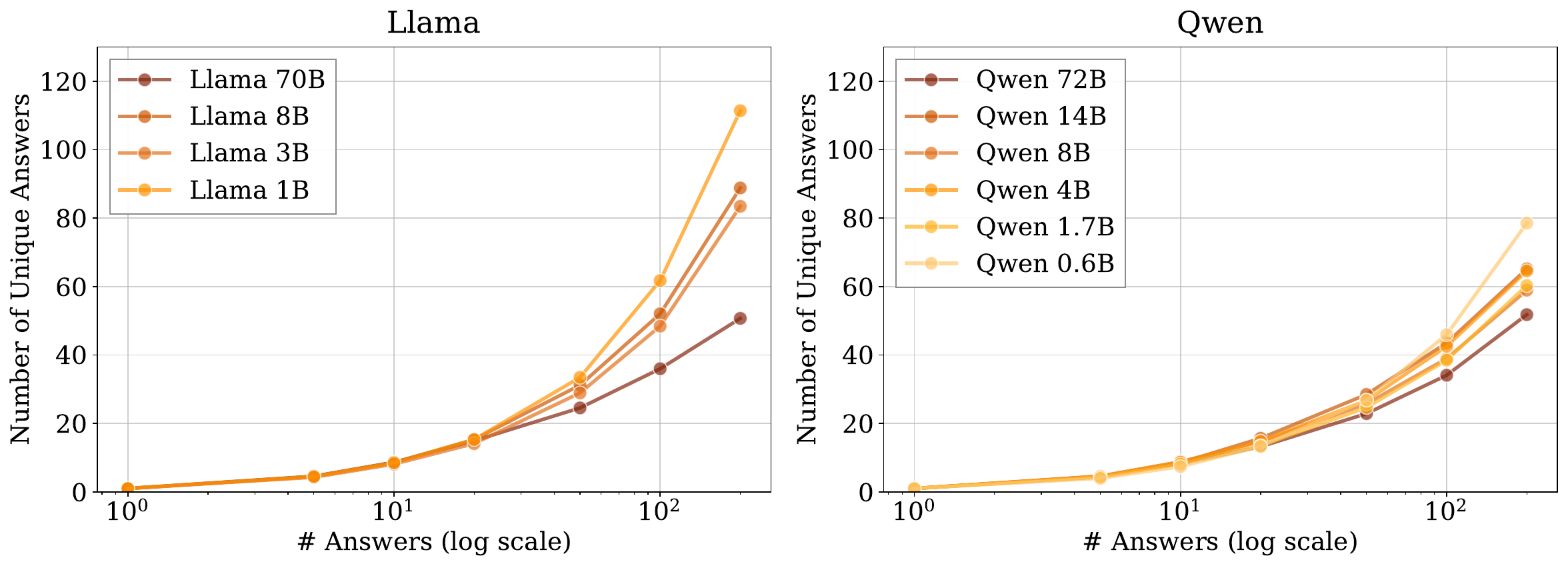}
    
    \caption{
    Smaller models generate more unique answers. As number of inferred answers increases, 
    Llama models have better output uniqueness than Qwen, but also have more within-family variance. }
    \label{fig:unqiue_answer} 
\end{figure}
\paragraph{How does model size affect diversity?}
According to Figure~\ref{fig:scaling}, most models' performances are pretty similar on \sq, except for the smallest model Qwen 0.6B). On \nb,  medium-sized models (Llama 8B and Qwen 14B) consistently have higher overall diversity than extremely large or small ones.  We hypothesize that these models balance answer distinctness and quality best, therefore achieving the highest diversity performance. Figure~\ref{fig:unqiue_answer} shows answer uniqueness is inversely proportional to the model sizes. And Figure~\ref{fig:tradeoff} shows answer quality is proportional to model size.
Finally, we also noticed that model rankings are largely unchanged regardless of the number of collected answers.

\end{document}